\documentclass[twocolumn]{article}

\usepackage[top=15truemm,bottom=20truemm,left=15truemm,right=15truemm]{geometry}

\usepackage{algorithm}
\usepackage{algorithmic}
\usepackage{amsmath}
\usepackage{amssymb}
\usepackage{amsthm}
\usepackage{array}
\usepackage{booktabs}
\usepackage{cite}
\usepackage{graphicx}
\usepackage{hyperref}
\usepackage{multirow}
\usepackage{physics}
\usepackage{url}

\newcommand{\eg}{\emph{e.g.}}
\newcommand{\ie}{\emph{i.e.}}
\newcommand{\etal}{\emph{et al.}}

\newtheorem{theorem}{Theorem}
\newtheorem{definition}{Definition}

\newcommand{\argmax}{\mathop{\rm arg~max}\limits}

\begin{document}

\title{Are DNNs fooled by extremely unrecognizable images?\footnotemark[1]}

\author{Soichiro Kumano\thanks{kumano@cvm.t.u-tokyo.ac.jp} \and Hiroshi Kera\thanks{kera@chiba-u.jp} \and Toshihiko Yamasaki\thanks{yamasaki@cvm.t.u-tokyo.ac.jp}}

\date{}

\maketitle

\begin{abstract}
  Fooling images are a potential threat to deep neural networks~(DNNs). These images are not recognizable to humans as natural objects, such as dogs and cats, but are misclassified by DNNs as natural-object classes with high confidence scores. Despite their original design concept, existing fooling images retain some features that are characteristic of the target objects if looked into closely. Hence, DNNs can react to these features. In this paper, we address the question of whether there can be fooling images with no characteristic pattern of natural objects locally or globally. As a minimal case, we introduce single-color images with a few pixels altered, called sparse fooling images~(SFIs). We first prove that SFIs always exist under mild conditions for linear and nonlinear models and reveal that complex models are more likely to be vulnerable to SFI attacks. With two SFI generation methods, we demonstrate that in deeper layers, SFIs end up with similar features to those of natural images, and consequently, fool DNNs successfully. Among other layers, we discovered that the max pooling layer causes the vulnerability against SFIs. The defense against SFIs and transferability are also discussed. This study highlights the new vulnerability of DNNs by introducing a novel class of images that distributes extremely far from natural images.
\end{abstract}

\footnotetext[1]{This work is partially financially supported by JSPS KAKENHI Grant Number JP19K22863.}

\begin{figure}[t]
\centering
\includegraphics[width=\linewidth]{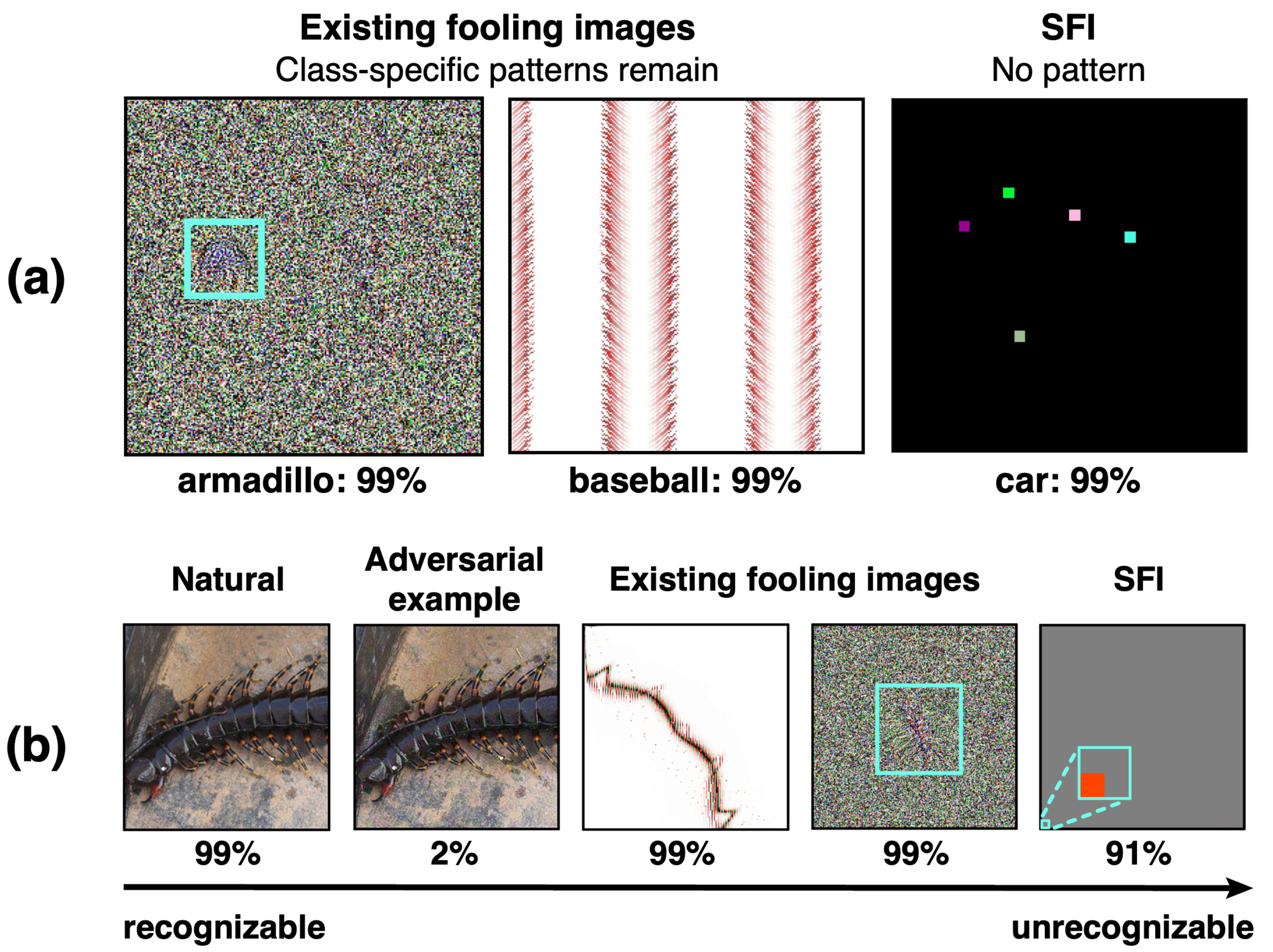}
   \caption{(a)~An example of images making DNNs behave strangely. Although fooling images are not recognizable to humans as natural images, the classifiers recognize them as natural objects. However, existing fooling images still retain some class-specific patterns~(carapace of armadillo highlighted by cyan square and baseball stitch). Our SFI, proposed as a new class of fooling images, does not have any features perceivable by humans locally or globally. (b)~"Centipede" images. Adversarial examples are similar to natural images and fooling images are not so. In particular, SFIs are extremely different from natural images.}
\label{fig:abst}
\end{figure}

\section{Introduction}
It is well known that deep neural networks~(DNNs) can be fooled by images far from those used for training~\cite{amodei2016concrete, hein2019relu, hendrycks17baseline, nguyen2015deep, sehwag2019better}. A typical example is fooling images~\cite{nguyen2015deep}. In contrast to notorious adversarial examples~\cite{szegedy2013intriguing}, which are subtly perturbed natural images, fooling images are designed to distribute extremely far from natural images and thus unrecognizable to humans. Nevertheless, fooling images are classified by DNNs into certain classes with high confidence scores~(Figure~\ref{fig:abst}). This reveals a crucial difference between human and DNN perception, highlighting the vulnerability of learning by DNNs on natural images. However, existing fooling images retain some features that are obviously specific to natural objects~(\eg, the armadillo's carapace-like pattern in Figure~\ref{fig:abst}(a)), implying that DNNs merely respond to the local features or global patterns of the images. Therefore, it remains unexplored whether \textit{completely} featureless images can fool DNNs.

In this paper, we address a fundamental question: \textit{Is it possible to fool DNNs with images that are extremely far from natural ones and thus unrecognizable to humans?} Answering this question is crucial for the safe and secure use of DNNs because if it is possible, then DNNs must be robust not only against perturbed images that are close to natural images~(\ie, adversarial images) but also against those that are extremely far from them, expanding the range of potential threats to DNNs to concern.

We answer the above question affirmatively---DNNs \textit{can} be fooled by extremely unrecognizable images---through the analysis of sparse fooling images~(SFIs), which we propose as a minimal case of such unrecognizable images. SFIs comprise a single-color image with a small number of altered pixels and have no human recognizable features, either globally or locally, unlike existing fooling images~(Figure~\ref{fig:abst}(a)). It is difficult to imagine that a small number of pixels~(\eg, 5 pixels in a 32$\times$32 image, approximately 0.5\% of the whole image) can significantly increase the confidence score on a certain class and change the classifier prediction. Nevertheless, we discovered that DNNs classify SFIs into classes of natural objects with high confidence scores~(Figure~\ref{fig:abst}(a)).

Through theoretical and empirical analysis, we reveal why and to what extent DNNs are vulnerable to SFIs. We first prove that SFIs always exist under mild conditions for linear functions, one-hidden-layer neural networks, and kernel methods. The theoretical conditions we derived suggest that as the number of altered pixels increases, SFIs are more likely to exist; namely, complex models are more vulnerable to SFIs.

For the empirical analysis using DNNs, we developed two methods of generating SFIs\footnote{Our code will be available on \url{https://github.com/s-kumano/sparse-fooling-image}.}. The first method is based on differential evolution~(DE)~\cite{das2010differential, storn1997differential} and applied to a class-targeted setting. The second method is a transformation from a natural image to an SFI and is applied to an image-targeted setting. Extensive experiments revealed that SFIs, which have no recognizable features at all, are classified into classes with high confidence scores by the complex DNNs~(\eg, VGG16~\cite{simonyan2014very} and ResNet50~\cite{he2016deep}). The reason for this can be found in the similarity of deep features between SFIs and natural images. We revealed that features of SFIs, which show a completely different distribution from that of natural images in the input space, become similar to those of natural images after passing through several layers of DNNs. To examine which layer makes DNNs vulnerable, we compared the four network architectures: multi-layer perceptron~(MLP), a convolutional neural network~(CNN) with and without max pooling layers~(denoted as CNN$_\text{MP}$ and CNN$_\text{None}$, respectively), and CapsNet~\cite{sabour2017dynamic}. We observed that SFIs fool CNN$_\text{MP}$ most successfully, \ie, max pooling leads to the vulnerability of DNNs. Max pooling discards spatial information and makes decisions based on just whether lower-level features exist, not how they are connected. As a defense against SFIs, we tested out-of-distribution~(OOD) detection, which tries to classify the SFIs into an additional class. After training with SFIs in the OOD framework, the trained model successfully detected the SFIs in the test dataset. However, it was still possible to generate another set of SFIs that successfully fools such a model. In addition, the trained model did not detect most SFIs with a base color that was different from those used during training. These results indicate the difficulty of detecting SFIs through OOD detection and suggest the inherent vulnerability of DNNs. Lastly, we show that SFIs generated for one model can transfer to another model. Thus, altered pixels of SFIs form some features that are unrecognizable but commonly learned by different models.

Our contributions are summarized as follows.

\begin{itemize}
    \item We reveal a new vulnerability of DNNs through SFIs. In contrast to adversarial examples and existing fooling images, SFIs have neither local nor global features, and distribute extremely far from natural images. Analysis on such unrecognizable images is important in terms of safe and security of DNNs.
    \item We first prove the existence of SFIs under mild conditions for three models~(Theorems~\ref{th:linear}--\ref{th:kernel}). The conditions imply that complex models are more vulnerable to SFIs.
    \item We propose DE- and transformation-based methods to generate SFIs for empirical studies, and demonstrate that the features of natural images and SFIs were similar in deep layers of DNNs. This can partially explain why SFIs are classified into certain classes with high confidence scores.
    \item By testing four different architectures~(MLP, CNN$_\text{MP}$, CNN$_\text{None}$, and CapsNet), we reveal the defect of max pooling.
    \item We find that OOD-based defense against SFIs is not sufficient, \ie, various SFIs can be generated to attack the defended model.
    \item We show that the SFIs transfer between models. This suggests that DNNs might learn common completely unrecognizable features to humans.
\end{itemize}

The results listed above all indicate the potential threat of images that are extremely far from natural images. While SFIs are an example of such potentially harmful images, there can be more variety of such making-no-sense images that fool DNNs. We believe that extensive studies on them will be crucial for the safe and secure applications of DNNs. 

\section{Related Work}
\label{sec:related}
\paragraph{Fooling Images and Out-of-Distribution Images.}
Nguyen \etal~proposed fooling images, which do not make sense to humans, but are classified by DNNs with high confidence scores into certain classes~(Figure~\ref{fig:abst}(a))~\cite{nguyen2015deep}. Some parts and the overall pattern of the fooling images have the characteristics of natural objects. The authors hypothesized that fooling images produce high-confidence predictions despite being far from the natural images in the class because they are typically far from the decision boundary and deep in a classification region. Hein \etal~proved that DNNs with a rectified linear unit~(ReLU) activation produce high-confidence predictions for samples sufficiently far from the training data~\cite{hein2019relu}. In their study, by relying on the fact that ReLU DNNs split the input space into linear regions~\cite{arora2016understanding} in most cases, it was shown that for any input sample $x$, the scaling $ax$ would fall into a consistent linear region when $a$ exceeds a certain value~(say, $a_0$). Within such a linear region, the ReLU DNN consistently classifies $ax$ into the same class for any $a\geq a_0$, and the probability~(softmax value) assigned to this class increases monotonically along $a$. In addition, Simonyan \etal~generated white noise-like images by computing the gradient of the class score with respect to the input image to confirm what the CNNs capture~\cite{simonyan2013deep}. These images are not natural images, but have some specific features similar to the target class. In this paper, we deal with completely unrecognizable images, \ie, images that do not have class-specific features locally or globally. The existence of our proposed images, SFIs, is proved theoretically and experimentally. Note that Hein \etal~theoretically proved that any image with extreme scaling can fool DNNs, but they assume unbounded pixel values. In contrast, with SFIs, we theoretically and experimentally demonstrated an example of extreme OOD with bounded pixel values. SFIs suggest the fundamental vulnerability and risk of DNNs that completely featureless images, even single-color images with a few pixels altered, can break the prediction of DNNs.

\paragraph{Adversarial Examples.}
A similar but distinct concept from fooling images is adversarial examples, which are natural images with small malicious perturbations~\cite{szegedy2013intriguing, carlini2017towards, chen2017ead, dong2018boosting, moosavi2016deepfool,eykholt2018robust, sharif2016accessorize, sharif2019general, thys2019fooling, goodfellow2014explaining, Croce_2019_ICCV, Modas_2019_CVPR, narodytska2016simple, papernot2016limitations, su2019one}. 
The small perturbations imply that adversarial examples are close to the original natural images. Since we address the question of whether it is possible to fool DNNs by images that are extremely far from natural ones, we claim that adversarial examples should be separated from SFIs, and thus they are not of our interest. In particular, one pixel attack~\cite{su2019one} might seem similar to our study in that it changes the predictions of DNNs with a change in the small number of pixels; however, its research objective is different from ours. The study of one pixel attack focuses on local perturbations of natural images, whereas our study aims to investigate the potential threat of images that are extremely far from natural images. Methodologically, one of our SFI generation methods is based on the one pixel attack framework. However, this cannot produce image-targeted generation. Hence, we also propose the other method that generates an SFI from a target natural image by inverting their deep features. Furthermore, in contrast to one pixel attack, SFIs are theoretically well-supported in terms of their existence for several linear and nonlinear models. 

\section{SFI Generation}
\label{sec:method}
An SFI comprises a single color image with a small number of altered pixels. In particular, we refer to an SFI with $k$ altered pixels as a $k$-SFI. For example, the SFIs in Figure~\ref{fig:SFI} are 5-SFIs. A formal definition of $k$-SFIs will be provided later~(cf.~Definition~\ref{def:SFI}).

We propose two methods for generating SFIs. The first method is a DE-based generation in the class-targeted setting, which works in a semi-black-box setting where only confidence scores are assumed to be accessible. The second method is a transformation-based generation in the image-targeted setting, which works in a white-box setting where all parameters and scores are assumed to be accessible. The second method is later used to highlight that even two significantly distant images can be interpreted similarly by DNNs. 

\subsection{SFI generation through differential evolution}
\label{sec:deMethod}
Following the framework proposed by Su \etal~\cite{su2019one}, we use DE to optimally select the position and intensity of pixels in the SFIs. The generation of a grayscale $k$-SFI is conducted as follows:

\begin{enumerate}
  \item Candidate solutions are first generated for $s$. Each candidate solution is a $k \times 3$ matrix. The row of the matrix is $(x,y,c)$, which represents the position $(x, y)$ and the intensity $c$ of the pixel to be altered\footnote{In the RGB color image, a five-dimensional vector $(x, y, R, G, B)$ is used.}. The position of a pixel is sampled from a uniform distribution of all pixel positions, and the intensity of a pixel is sampled from the normal distribution $N\left(\frac{m+n}{2},\left(\frac{m-n}{2}\right)^2\right)$ of the initial step, where $[m, n]$ is the predefined range of pixel intensity. To maintain the pixel intensity $c$ in the range $[m, n]$, $c$ is mapped to $n$ if $c$ is higher than $n$, and to $m$ if $c$ is lower than $m$.
  \item Repeat the following steps for $t = 1, 2, \ldots, T$.
  \begin{enumerate}
    \item A tentative group of candidate solutions $\{v_{i, t+1}\},~(i=1, 2, \ldots, s)$ is generated from the current one $\{x_{i, t}\},~(i=1, 2, \ldots, s)$ as $v_{i, t+1}=x_{r_1, t}+F\cdot\left(x_{r_2, t}-x_{r_3, t}\right)$. Then, three non-identical indices of the candidate solutions, $r_1$, $r_2$, and $r_3$ are randomly selected. In addition, $F$ is the scaling factor (here, $F = 0.5$).
    \item The original group of candidate solutions is updated by replacing $x_{i,t}$ with $v_{i,t+1}$ when $v_{i,t+1}$ has a higher confidence score on the target class; that is, $x_{i, t+1}=v_{i, t+1}$ if $f\left(x_{i, t}\right) < f\left(v_{i,t+1}\right)$ and $x_{i, t+1}=x_{i, t}$, otherwise, where $f(\cdot)$ denotes the confidence score on the target class of the SFI generated from a candidate solution and the predesignated base intensity $g$.
  \end{enumerate}
  \item After $T$ iterations, an SFI is generated from the candidate solution with the highest confidence score on the target class in the current group of candidate solutions.
\end{enumerate}

\subsection{SFI generation through deep feature inversion}
\label{sec:invertMethod}
To generate image-targeted SFIs, we impose two requirements on an SFI:~(i)~the features of the SFI through the model are similar to those of a natural image,~and~(ii)~almost all the pixels of the SFI have the same value, which can be formulated as
\begin{align}
\small
\vb*{\alpha}^{\ast} = \arg\min_{\vb*{\alpha}} \frac{||\Phi(z_b\vb{1}+\vb*{\alpha})-\Phi(\vb{x}_0)||_2^2}{||\Phi(\vb{x}_0)||_2^2} + \lambda_{\mathrm{sparse}} ||\vb*{\alpha}||_1 ,
\normalsize
\label{eq:natural2pixel}
\end{align}
where $\| \cdot \|_{p}$ is the $\ell_p$ norm of a vector, $\Phi(\cdot)$ is the output of the DNN or its intermediate layer, $\lambda_{\mathrm{sparse}}$ is the weight for the regularization, and $z_b\vb{1}$ is a single color image~(\ie, $z_b$ is zero to generate black-based SFIs). The first term represents the similarity between the outputs of a natural image $\vb{x}_0$ and an SFI $z_b\vb{1}+\vb*{\alpha}$. Natural images and SFIs generated by DE have a similar distribution at the deep layer of the network. Here, the first term explicitly requires a similarity between the given natural image and the generated SFI, allowing the SFI to fool the model more effectively. The second term is a regularization that encourages the sparsity of $\vb*{\alpha}$~(\ie, this method does not set the value of $k$). This is similar to the method proposed by Mahendran \etal~\cite{mahendran2015understanding}. They used different regularization terms, which suppress extreme pixels and image flicker, to reconstruct the input image from the output of the CNNs.

\begin{figure}[t]
\centering
\includegraphics[width=\linewidth]{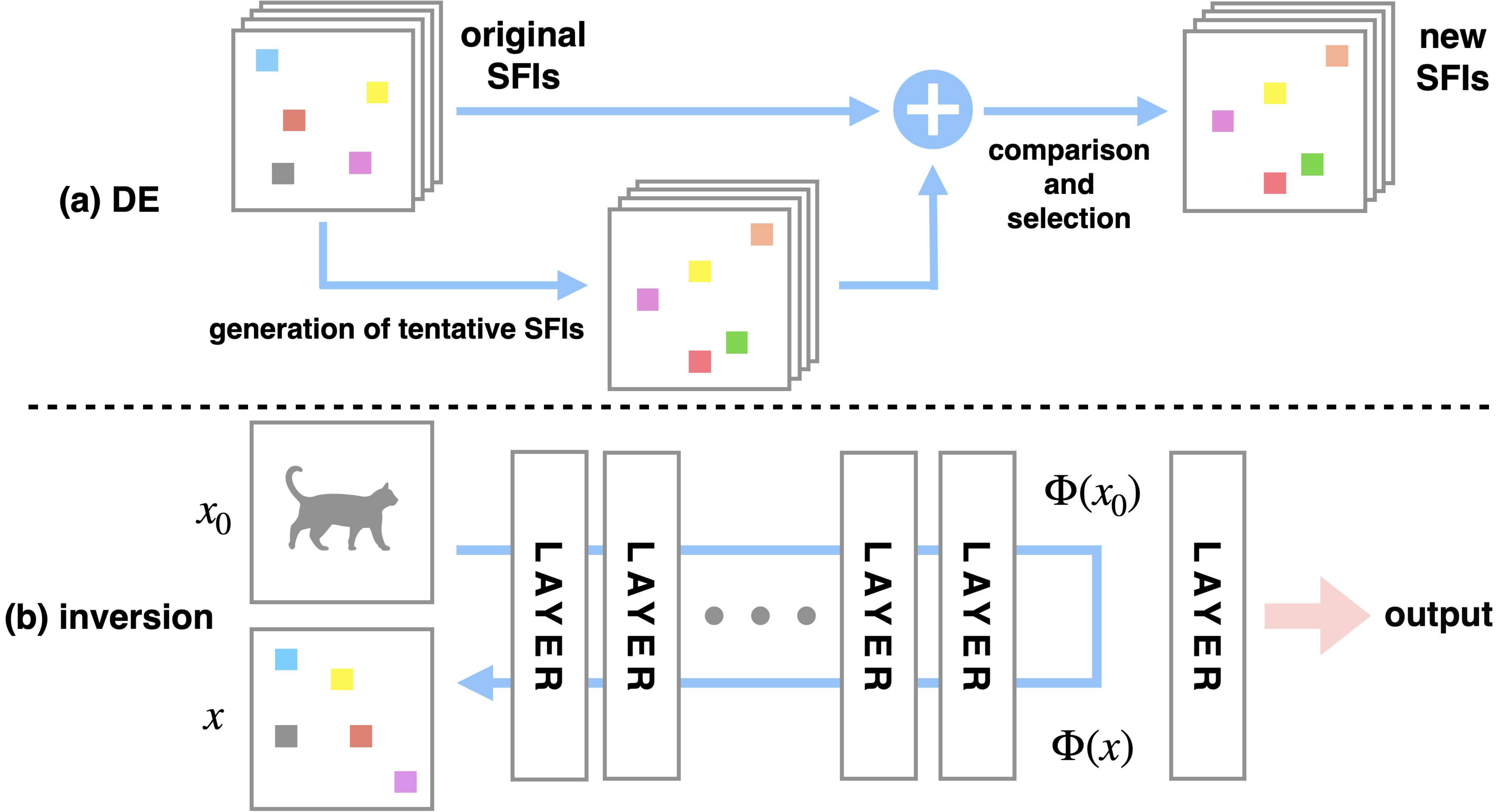}
   \caption{Two methods proposed for generating SFIs. (a)~Schematic diagram of DE used to generate SFIs. The original group of candidate solutions is updated by comparing the tentative one with the confidence scores on the target class. By repeating this step, we obtain an SFI. (b)~Schematic diagram of deep feature inversion used to generate an SFI using the deep features of a natural image. We search for sparse images where $\Phi(x_0)$ and $\Phi(x)$ are similar using backpropagation.}
\label{fig:method}
\end{figure}

\section{Existence of SFIs}
\label{sec:exi}
Formally, a $k$-SFI is defined as follows\footnote{$k$-SFIs for a vector and a three-dimensional tensor~(color image) are defined similarly.}:
\begin{definition}
\label{def:SFI}
An image $\vb{z} \in[r, R]^{H\times W}$ is called a $k$-SFI if it can be represented by $\vb{z} = z_b\vb{1}_{H, W} + \vb*{\alpha}$, where $z_b\in[r, R]$, $\vb{1}_{H,W}\in\mathbb{R}^{H\times W}$ is an all-one matrix, and $\vb*{\alpha}\in\mathbb{R}^{H\times W}$ is a matrix with at most $k$ nonzero entries.
\end{definition}

We theoretically prove the existence of SFIs for simple models. Specifically, we discuss conditions for the existence of SFIs for a linear function, a one-hidden-layer neural network, and a kernel method. For simplicity, we consider a binary classification~(between classes $C_1$ and $C_2$), and without a loss of generality, an attack on class $C_1$. A natural image $\vb{x}$ is classified as $C_1$ if the confidence score is $p_1(\vb{x})>\frac{1}{2}$. Now, the question is whether there exists a bounded SFI $\vb{z}\in[r,R]^D$ misclassified with a high confidence score by the binary classifier. Specifically, we consider the conditions on the weight $\vb{W}$ such that there exists $\vb{z}$ with $p_1(\vb{z})\geq M$, where $M\in[0,1]$.

\subsection{Linear function}
First, we consider the conditions of $W\in\mathbb{R}^{2\times D}$ of a linear function $\vb{y}(\vb{x})=\vb{W}\vb{x}$ for the existence of SFIs. We can design an optimal $k$-SFI, and a sufficient and necessary condition is given as follows: 

\begin{theorem}
  Let $\vb{W}=\mqty[\vb{w}_1&\vb{w}_2]^\top\in\mathbb{R}^{2\times D}$ be the weight matrix of the linear function $\vb{y}(\vb{z})=\vb{W}\vb{z}$ for the binary classification of classes $C_1$ and $C_2$. Let $\vb*{\delta}_w=(\delta_{w,1},\ldots,\delta_{w,D})^\top=\vb{w}_1-\vb{w}_2$. For any $\vb{W}$, there exists a $k$-SFI $\vb{z}\in[r,R]^D$ with a confidence score of at least $M\in[0,1]$ on $C_1$ if and only if the following holds: If $\vb*{\delta}_w^\top\vb{1}>0$, then
  \begin{align}
    \ln\qty(\frac{M}{1-M})\leq R\vb*{\delta}_w^\top\vb{1}+\qty(r-R)\sum_{i\in I_{-}(k)}\delta_{w,i},
  \label{eq:linear0}
  \end{align}
where $I_{-}(k)\subset\{1,\ldots,D\}$ is the index set of~(at most) $k$ smallest negative entries in $\vb*{\delta}_w$. Otherwise,
  \begin{align}
    \ln\qty(\frac{M}{1-M})\leq r\vb*{\delta}_w^\top\vb{1}+\qty(R-r)\sum_{i\in I_{+}(k)}\delta_{w,i},
  \label{eq:linear1}
  \end{align}
where $I_{+}(k)\subset\{1,\ldots,D\}$ is the index set of~(at most) $k$ largest positive entries in $\vb*{\delta}_w$.
  \label{th:linear}
\end{theorem}

\begin{proof}
Let us denote an SFI by $\vb{z}=z_b\vb{1}+\vb*{\alpha}\in[r,R]^D$. The confidence score of the SFI on $C_1$ is as follows:
\begin{align}
    p_1(\vb{z})=\frac{\exp(\vb{w}_1^\top\vb{z})}{\exp(\vb{w}_1^\top\vb{z})+\exp(\vb{w}_2^\top\vb{z})}.
\end{align}
Thus, there exists an SFI with $p_1(\vb{z})\geq M$ when the following holds:
\begin{align}
    \frac{\exp(\vb{w}_1^\top\vb{z})}{\exp(\vb{w}_1^\top\vb{z})+\exp(\vb{w}_2^\top\vb{z})} \geq M.
    \label{eq:per1}
\end{align}
(\ref{eq:per1}) can be simplified to
\begin{align}
    \ln\qty(\frac{M}{1-M})\leq
    \sum_{d=1}^D\delta_{w,d}(z_b+\alpha_d)
    =:f(z_b,\vb*{\alpha}).
    \label{eq:per2}
\end{align}
Here, we consider maximizing $f(z_b,\vb*{\alpha})$. If $\delta_{w,d}>0$, $f(z_b,\vb*{\alpha})$ can be maximized by $z_b+\alpha_d=R$ because pixel intensities of SFIs should range in $[r,R]$. Similarly, if $\delta_{w,d}<0$, $z_b+\alpha_d=r$ can maximize $f(z_b,\vb*{\alpha})$. Thus, if $\vb*{\alpha}$ had no constrained condition, we would choose
\begin{align}
    z_b+\alpha_d = 
    \begin{cases}
        R, & (\delta_{w,d}>0) \\
        r, & (\delta_{w,d}\leq0)
    \end{cases}
    \label{eq:per3}
\end{align}
to maximize $f(z_b,\vb*{\alpha})$. However, $\vb*{\alpha}$ is restricted to have only $k$ nonzero entries. We can select the nonzero entries of $\vb*{\alpha}$ as follows: if $\vb*{\delta}_w^\top\vb{1}>0$, we define $I_{-}(k)$ as the index set of~(at most) $k$ smallest negative entries in $\vb*{\delta}_w$. Then, for $i\in I_{-}(k)$, $\alpha_i$ is set to $r-R$~(thus, $z_b=R$). If $\vb*{\delta}_w^\top\vb{1}\leq0$, we define $I_{+}(k)$ as the index set of~(at most) $k$ largest positive entries in $\vb*{\delta}_w$. Then, for $i\in I_{+}(k)$, $\alpha_i$ is set to $R-r$~(thus, $z_b=r$). These maximizes $f(z_b, \vb*{\alpha})$, \textit{i.e.}, 
\footnotesize
\begin{align}
    \max f(z_b,\vb*{\alpha}) =
    \begin{cases}
        R\vb*{\delta}_w^\top\vb{1}+\qty(r-R)\sum_{i\in I_{-}(k)}\delta_{w,i}, & (\vb*{\delta}_{w}^\top\vb{1}>0) \\
        r\vb*{\delta}_w^\top\vb{1}+\qty(R-r)\sum_{i\in I_{+}(k)}\delta_{w,i}, & (\vb*{\delta}_{w}^\top\vb{1}\leq0)
    \end{cases}.
    \label{eq:per4}
\end{align}
\normalsize
In conclusion, there exists a $k$-SFI with a confidence score of at least $M$ on $C_1$ if and only if
\footnotesize
\begin{align}
    \ln\qty(\frac{M}{1-M})\leq
    \begin{cases}
        R\vb*{\delta}_w^\top\vb{1}+\qty(r-R)\sum_{i\in I_{-}(k)}\delta_{w,i}, & (\vb*{\delta}_{w}^\top\vb{1}>0) \\
        r\vb*{\delta}_w^\top\vb{1}+\qty(R-r)\sum_{i\in I_{+}(k)}\delta_{w,i}, & (\vb*{\delta}_{w}^\top\vb{1}\leq0)
    \end{cases}.
    \label{eq:per5}
\end{align}
\normalsize
\end{proof}

For instance, when $r = 0$, $R = 255$, $M = 0.9$, and $\vb*{\delta}_{w}^\top\vb{1}\leq0$, (\ref{eq:linear1}) becomes $\ln\qty(9) \leq 2.2 \leq 255\sum_{i\in I_{+}(k)}\delta_{w,i}$. This condition can be easily satisfied because $\vb{w_1}$ and $\vb{w_2}$ are not similar, and thus natural data are well classified and $\vb*{\delta}_w$ can have sufficiently large positive entries~(\ie, a large $\sum_{i\in I_{+}(k)}\delta_{w,i}$). Hence, SFIs exist for almost all linear classifiers. Our results also reveal that as the range of pixel intensity, $R-r$, increases, satisfying this inequality becomes easier. This observation is consistent with~\cite{hein2019relu}, where an input with sufficiently large pixel intensities can fool the classifiers. 

\subsection{One-hidden-layer neural network}
As a nonlinear model, we consider a one-hidden-layer neural network with weight matrices $\vb{U}$ of the hidden layer and $\vb{V}$ of the output layer. Again, the bias was reduced. The conditions on $\vb{U}$ and $\vb{V}$, such that an SFI $\vb{z}$ with confidence score $p_1(\vb{z})\geq M$ exists, are as follows:

\begin{theorem}
    Let $\vb{U}=\mqty[\vb{u}_1&\cdots&\vb{u}_{D_h}]^\top\in\mathbb{R}^{D_h\times D}$ be the weight of a hidden layer and $\vb{V}=\mqty[\vb{v}_1&\vb{v}_2]^\top\in\mathbb{R}^{2\times D_h}$ be the weight of an output layer of a one-hidden-layer neural network $\vb{y}(\vb{x})=\vb{V}(\sigma\odot(\vb{U}\vb{z}))$ for the binary classification between classes $C_1$ and $C_2$, where $D_h$ is dimension of the hidden layer and $\sigma \odot (\cdot)$ applies a ReLU function to each entry of a vector. Let $\vb*{\delta}_v=(\delta_{v,1},\ldots,\delta_{v,D_h})^\top=\vb{v}_1-\vb{v}_2$. For any $\vb{U}$, $\vb{V}$, and $M\in[0,1]$, there exists a $k$-SFI $\vb{z}\in[0,R]^D$ with a confidence score of at least $M$ on $C_1$ if
    \begin{align}
        \ln\qty(\frac{M}{1-M})\leq
        R\sum_{d_h=1}^{D_h}\delta_{v,d_h}\sigma\odot\qty(\sum_{i\in I(k)}\vb{u}_{d_h,i}),
        \label{eq:hidden}
    \end{align}
    where $I(k)$ is determined by Algorithm~\ref{algo:onehidden}.
    \label{th:hidden}
\end{theorem}

\begin{proof}
Let us denote an SFI by $\vb{z}=z_b\vb{1}+\vb*{\alpha}\in[0,R]^D$. The confidence score of the SFI on $C_1$ is as follows:
\begin{align}
    p_1(\vb{z})
    =\frac{
        \exp(\vb{v}_1^\top \sigma \odot (\vb{U}\vb{z}))
    }{
        \exp(\vb{v}_1^\top \sigma \odot (\vb{U}\vb{z}))
        +\exp(\vb{v}_2^\top \sigma \odot (\vb{U}\vb{z}))
    }.
\end{align}
Thus, there exists an SFI with $z_{b}=0$ and $p_1(\vb{z})\geq M$ when the following holds:
\begin{align}
    \frac{
        \exp(\vb{v}_1^\top \sigma \odot (\vb{U}\vb*{\alpha}))
    }{
        \exp(\vb{v}_1^\top \sigma \odot (\vb{U}\vb*{\alpha}))
        +\exp(\vb{v}_2^\top \sigma \odot (\vb{U}\vb*{\alpha}))
    } \geq M.
    \label{eq:hid1}
\end{align}
(\ref{eq:hid1}) can be simplified to
\begin{align}
    \ln\qty(\frac{M}{1-M})&\leq
    \sum_{d_h=1}^{D_h}\delta_{v,d_h}\sigma \odot \qty(\sum_{d=1}^D\vb{u}_{d_h,d}\alpha_{d})\\
    &=:h(\vb*{\alpha}).
    \label{eq:hid2}
\end{align}
We consider making $h(\vb*{\alpha})$ as large as possible. We only need to choose the index set of altered pixels $I(k)\subset\{1,\ldots,D\}$ by assuming $\alpha_d\in\qty{0,R}$ for any $d\in\qty{1,\ldots,D}$; that is, we consider making
\begin{align}
    h'(I(k)) &:= h\qty(\vb*{\alpha}\in\qty{0,R}^D)\\
    &= R\sum_{d_h=1}^{D_h}\delta_{v,d_h}\sigma \odot\qty(\sum_{i\in I(k)}u_{d_h,i})
    \label{eq:hid3}
\end{align}
\normalsize
as large as possible rather than $h(\vb*{\alpha})$. Here, $I(k)$ is determined by Algorithm~\ref{algo:onehidden}. Algorithm~\ref{algo:onehidden} tries to maximize $h'(I(k))$ as large as possible by determining an element of $I$ sequentially. Although in this case, $I(k)$ does not maximize $h'(I(k))$, $h'(I(k))$ monotonically increases with respect to~$k$. In conclusion, there exists a $k$-SFI with a confidence score of at least $M$ on $C_1$ if
\begin{align}
    \ln\qty(\frac{M}{1-M})\leq R\sum_{d_h=1}^{D_h}\delta_{v,d_h}\sigma \odot\qty(\sum_{i\in I(k)}u_{d_h,i}).
    \label{eq:hid4}
\end{align}
\end{proof}

\begin{algorithm}[t]
    \caption{Algorithm for determination of $I$ on a one-hidden-layer neural network}
    \begin{algorithmic}
        \STATE \bf{Input:} $k\in\mathbb{N}$, $\vb*{\delta}_v\in\mathbb{R}^{D_h}$, $\vb{U}\in\mathbb{R}^{D_h\times D}$
        \STATE \bf{Output:} $I$
        \STATE $I\leftarrow\varnothing$
        \STATE $\mathcal{D}\leftarrow\qty{1,\ldots,D}$
        \STATE $h\leftarrow0$
        \WHILE{$|I|<n$}
            \STATE $i^{\ast}\leftarrow\argmax_{i\in\mathcal{D}}\sum_{d_h=1}^{D_h}\delta_{v,d_h}\sigma\odot\qty(\sum_{j\in I\cup\qty{i}}u_{d_h,j})$
            \STATE $h_{\text{new}}\leftarrow\sum_{d_h=1}^{D_h}\delta_{v,d_h}\sigma\odot\qty(\sum_{j\in I\cup\qty{i^{\ast}}}u_{d_h,j})$
            \IF{$h_{\text{new}}>h$}
                \STATE $I\leftarrow I\cup\qty{i^{\ast}}$
                \STATE $\mathcal{D}\leftarrow\mathcal{D}\backslash\qty{i^{\ast}}$
                \STATE $h\leftarrow h_{\text{new}}$
            \ELSE
                \STATE \textbf{break}
            \ENDIF
        \ENDWHILE
        \STATE \bf{return} $I$
    \end{algorithmic}
    \label{algo:onehidden}
\end{algorithm}

\subsection{Kernel method}
As a nonlinear model, we also consider a kernel method. A kernel method transforms the data in raw representation into feature vector representations, and classifies them in high-dimension space. Kernel method computes a weighted sum of similarities
\begin{align}
    \sum_{n=1}^{N}s_{C,n}K(\vb{x}_{C, n}, \vb{x}'),
    \label{eq:kerbefore}
\end{align}
where $N$ is the number of data in a class, $\vb{s}_C$ is the weight vector, $\vb{x}_{C, n}$ is the $n$-th data in class $C\in\qty{C_1, C_2}$, $\vb{x}'$ is an unlabeled input, and $K$ is the kernel function. Also, by rearranging~(\ref{eq:kerbefore}), the following equation is obtained and weight vector in feature space $\vb{t}_C$ is defined:
\begin{align}
    \sum_{n=1}^{N}\vb{s}_{C,n}K(\vb{x}_{C, n}, \vb{x}')&=\qty(\sum_{n=1}^{N}\vb{s}_{C,n}\varphi(\vb{x}_{C, n}))^\top \varphi(\vb{x}')\\
    &=:\vb{t}_C^\top\varphi(\vb{x}'),
    \label{eq:kar}
\end{align}
\normalsize
where $K(\vb{x}_1, \vb{x}_2)=\varphi(\vb{x}_1)^\top\varphi(\vb{x}_2)$ and $\varphi(\cdot)$ is the feature vector. Here, let $K$ be homogeneous polynomial kernel~(HPK) of degree two $K(\vb{x}_1,\vb{x}_2)=(\vb{x}_1^\top\vb{x}_2)^2$. Let SFI be $\vb{z}=z_b\vb{1}+\vb*{\alpha}\in[0,R]^D$. The conditions on the weight vector in feature space $\vb{T}=\mqty[\vb{t}_1 & \vb{t}_2]^\top$ such that confidence score $p_1(\vb{z})\geq M$ are as follows:

\begin{theorem}
    Let $\vb{S}=\mqty[\vb{s}_1 & \vb{s}_2]^\top\in\mathbb{R}^{2\times N}$ be the weight matrix of a kernel method. $\vb{T}=\mqty[\vb{t}_1 & \vb{t}_2]^\top$ is defined as~(\ref{eq:kar}). We assume HPK of degree two and
    \footnotesize
    \begin{align}
        \phi(\vb{x})=(x_1^2,\ldots,x_D^2,\sqrt{2}x_1x_2,\sqrt{2}x_1x_3,\ldots,\sqrt{2}x_{D-1}x_D).
    \end{align}
    \normalsize
    Let $\vb*{\delta}_t=(\delta_{t,1},\ldots,\delta_{t,D})^\top=\vb{t}_1-\vb{t}_2$. For any $\vb{T}$ and $M\in[0,1]$, there exists an $k$-SFI $\vb{z}\in[0,R]^D$ with a confidence score of at least $M$ on $C_1$ if
    \footnotesize
    \begin{align}
        \ln\qty(\frac{M}{1-M})\leq
        R^2\sum_{i\in I(k)}\qty(
        \delta_{t,i}
        +\sqrt{2}\sum_{j\in I(k)\backslash\qty{1,\ldots,i}}\delta_{t,D+i+j-2}
        ),
        \label{eq:kernel}
    \end{align}
    \normalsize
    where $I(k)\subset\qty{1,\ldots,D}$ is determined by Algorithm~\ref{algo:kernel}.
    \label{th:kernel}
\end{theorem}

\begin{proof}
Let us denote an SFI by $\vb{z}=z_b\vb{1}+\vb*{\alpha}\in[0,R]^D$. The confidence score of the SFI on $C_1$ is as follows:
\begin{align}
    p_1(\vb{z})=\frac{\exp(\vb{t}_1^\top\vb{z})}{\exp(\vb{t}_1^\top\vb{z})+\exp(\vb{t}_2^\top\vb{z})}.
\end{align}
Thus, there exists an SFI with $z_{b}=0$ and $p_1(\vb{z})\geq M$ when the following holds:
\begin{align}
    \frac{\exp(\vb{t}_1^\top\vb*{\alpha})}{\exp(\vb{t}_1^\top\vb*{\alpha})+\exp(\vb{t}_2^\top\vb*{\alpha})} \geq M.
    \label{eq:ker1}
\end{align}
(\ref{eq:ker1}) can be simplified to
\footnotesize
\begin{align}
    \ln\qty(\frac{M}{1-M})&\leq
    \sum_{i=1}^D\alpha_i\qty(
        \delta_{t,i}\alpha_i+\sqrt{2}\sum^D_{j=i+1}\delta_{t,D+i+j-2}\alpha_j
    )\\
    &=:g(\vb*{\alpha}).
    \label{eq:ker2}
\end{align}
\normalsize
We consider making $g(\vb*{\alpha})$ as large as possible. We only need to choose the index set of altered pixels $I(k)\subset\{1,\ldots,D\}$ by assuming $\alpha_d\in\qty{0,R}$ for any $d\in\qty{1,\ldots,D}$; that is, we consider making
\footnotesize
\begin{align}
    g'(I(k))&:= g\qty(\vb*{\alpha}\in\qty{0,R}^D)\\
    &= R^2\sum_{i\in I(k)}\qty(
        \delta_{t,i}
        +\sqrt{2}\sum_{j\in I(k)\backslash\qty{1,\ldots,i}}\delta_{t,D+i+j-2}
    )
    \label{eq:ker3}
\end{align}
\normalsize
as large as possible rather than $g(\vb*{\alpha})$. Here, $I(k)$ is determined by Algorithm~\ref{algo:kernel}. Algorithm~\ref{algo:kernel} tries to maximize $h'(I(k))$ as large as possible by determining an element of $I$ sequentially. Although in this case, $I(k)$ does not maximize $g'(I(k))$, $g'(I(k))$ monotonically increases with respect to~$k$. In conclusion, there are exists an $k$-SFI with a confidence score of at least $M$ on $C_1$ if
\footnotesize
\begin{align}
    \ln\qty(\frac{M}{1-M})\leq R^2\sum_{i\in I(k)}\qty(
        \delta_{t,i}
        +\sqrt{2}\sum_{j\in I(k)\backslash\qty{1,\ldots,i}}\delta_{t,D+i+j-2}
    ).
    \label{eq:ker4}
\end{align}
\normalsize
\end{proof}

\begin{algorithm}[t]
    \caption{Algorithm for determination of $I$ on a kernel method}
    \begin{algorithmic}
        \STATE \bf{Input:} $k\in\mathbb{N}$, $\vb*{\delta}_t\in\mathbb{R}^D$
        \STATE \bf{Output:} $I$
        \STATE $I\leftarrow\varnothing$
        \STATE $\mathcal{D}\leftarrow\qty{1,\ldots,D}$
        \STATE $h\leftarrow0$
        \WHILE{$|I|<n$}
            \STATE $i^\ast\leftarrow\argmax_{i\in\mathcal{D}}\sum_{j\in I\cup\qty{i}}(\delta_{t,j}$
            \STATE $~\qquad+\sqrt{2}\sum_{k\in I\cup\qty{i}\backslash\qty{1,\ldots,j}}\delta_{t,D+j+k-2})$
            \STATE $h_{\text{new}}\leftarrow\sum_{j\in I\cup\qty{i^\ast}}(\delta_{t,j}$
            \STATE $~\qquad+\sqrt{2}\sum_{k\in I\cup\qty{i^\ast}\backslash\qty{1,\ldots,j}}\delta_{t,D+j+k-2})$
            \IF{$h_{\text{new}}>h$}
                \STATE $I\leftarrow I\cup\qty{i^\ast}$
                \STATE $\mathcal{D}\leftarrow\mathcal{D}\backslash\qty{i^\ast}$
                \STATE $h\leftarrow h_{\text{new}}$
            \ELSE
                \STATE \bf{break}
            \ENDIF
        \ENDWHILE
        \STATE \bf{return} $I$
    \end{algorithmic}
    \label{algo:kernel}
\end{algorithm}

\subsection{The effect of the number of altered pixels}
We discuss the effect of the number of altered pixels, $k$, in $k$-SFI attacks. The left-hand side of inequalities~(\ref{eq:linear0}) and~(\ref{eq:linear1}) increase at $O(\ln(M))$ for $M\in[0,1)$, whereas the right-hand side increases at $O(k)$ for $k\ll D$. Therefore, (\ref{eq:linear0}) and (\ref{eq:linear1}) are easily satisfied for a high confidence score $M$ by increasing $k$ slightly. In the one-hidden-layer neural network and a kernel method, we can obtain $O(D_hk)$ and $O(k^2)$, respectively, as the effect of the number of altered pixels.

\begin{proof}
In linear function~(cf. Theorem~1), if $\vb*{\delta}_w^\top\vb{1}>0$ and $\qty|\qty{i \mid \delta_{w,i}<0}|\gg k$, from~(\ref{eq:linear0}), we obtain
\begin{align}
    \small
    R\vb*{\delta}_w^\top\vb{1}+\qty(r-R)\sum_{i\in I_{-}(k)}\delta_{w,i}&\geq R\delta_w^\top\vb{1}+k\qty(r-R)\Delta_w\\
    \normalsize
    &=O(k),
\end{align}
where $\Delta_w=\max_{i\in I_{-}(k)}\delta_{w,i}$. Also, if $\vb*{\delta}_w^\top\vb{1}\leq0$ and $\qty|\qty{i \mid \delta_{w,i}>0}|\gg k$, from~(\ref{eq:linear1}), we obtain
\begin{align}
    r\vb*{\delta}_w^\top\vb{1}+\qty(R-r)\sum_{i\in I_{+}(k)}\delta_{w,i}
    &\geq r\delta_w^\top\vb{1}+k\qty(R-r)\Delta_w\\
    &= O(k),
\end{align}
where $\Delta_w=\min_{i\in I_{+}(k)}\delta_{w,i}$. Therefore, the vulnerability of a linear function against $k$-SFIs increases at $O(k)$. 
\end{proof}

\begin{proof}
In one-hidden-layer neural network~(cf. Theorem~2), if all entries of $\vb*{\delta_{v}}$ and $u_{d_h,i}$ are positive, where $d_h\in\qty{1,\ldots,D_h}$ and $i\in I$, from~(\ref{eq:hidden}), we obtain
\begin{align}
    R\sum_{d_h=1}^{D_h}\delta_{v,d_h}\sigma\odot\qty(\sum_{i\in I}u_{d_h,i})&\geq RD_hk\Delta_{v}\mu\\
    &= O(D_hk),
\end{align}
where $\Delta_{v}=\min_{d_h\in\qty{1,\ldots,D_h}}\delta_{v,d_h}$ and \\$\mu=\min_{d_h\in\qty{1,\ldots,D_h}, i\in I}u_{d_h,i}$. Therefore, the vulnerability of a one-hidden-layer neural network against $k$-SFIs increases at $O(D_hk)$.
\end{proof}

\begin{proof}
In the kernel method with HPK of degree 2~(cf. Theorem~\ref{th:kernel}), if all entries of $\vb*{\delta_{t}}$ are positive, from~(\ref{eq:kernel}), we obtain
\footnotesize
\begin{align}
    R^2\sum_{i\in I(k)}\qty(
        \delta_{t,i}
        +\sqrt{2}\sum_{j\in I(k)\backslash\qty{1,\ldots,i}}\delta_{t,D+i+j-2}
    )&\geq \frac{\sqrt{2}R^2k^2\Delta_t}{2}\\
    &= O(k^2),
\end{align}
\normalsize
where $\Delta_t=\min_{i\in I(k),j\in I(k)/\qty{1,\ldots,i}}\delta_{t,D+i+j-2}$. Therefore, the vulnerability of a kernel method against $k$-SFIs increases at $O(k^2)$.
\end{proof}

These results indicate that complex models are more likely to be vulnerable to SFIs. Although the analysis here is restricted to simple linear and nonlinear models, our results on these models theoretically support the empirical observations described in the Section~\ref{sec:result} in that SFIs can be found for various classes and datasets.

\section{Results}
\label{sec:result}

\subsection{Detailed settings}
\label{sec:detailedSettings}
\begin{figure}[t]
    \centering
    \includegraphics[width=\linewidth]{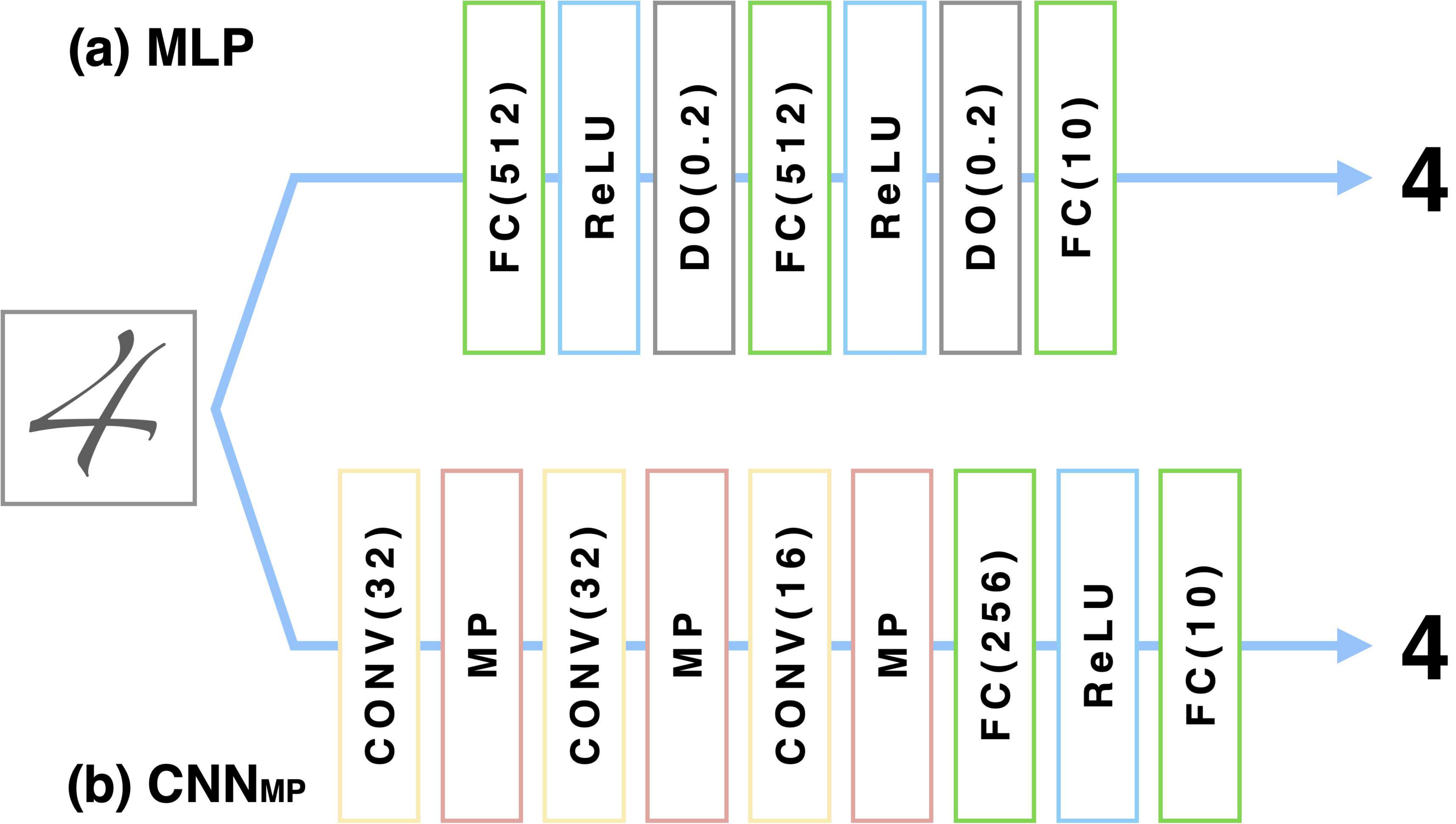}
    \caption{MLP and CNN architectures. FC and the number in parentheses refer to a fully connected layer and output dimensions, respectively. DO and the number in parentheses refer to a dropout layer and the probability of dropping neurons, respectively. CONV and the number in parentheses refer to a convolutional layer~(kernel size=$3\times3$, stride=$1$) with zero-padding of 1~pixel and output channels, respectively. MP refers to a max pooling layer.}
    \label{fig:archi}
\end{figure}

Here, we describe the experimental settings. Table~\ref{tab:accuracy} shows the top-1 accuracy of each architecture. The architectures on MNIST and FashionMNIST are trained using Adam~\cite{kingma2014adam} with learning rate=$0.001$. Those on CIFAR10~\cite{krizhevsky2009learning} are trained using stochastic gradient descent~(SGD) with learning rate=$0.01$, momentum=$0.9$, and weight decay=$5\times10^{-4}$. That on ImageNet~\cite{deng2009imagenet} is the Pytorch pretrained model~\cite{paszke2019pytorch}. Figure~\ref{fig:archi} shows detailed architectures of MLP and CNN. Dropout layers of MLP are used only during training.

\begin{table}[t]
  \caption{Top-1 accuracy~(\%) on natural test dataset.}
  \centering
  \begin{tabular}{@{}ccc@{}}
    \toprule
    dataset & architecture & accuracy~(top-1) \\ \midrule
    MNIST & MLP & 97 \\
    & CNN & 98 \\
    & CapsNet & 98 \\ \midrule
    FashionMNIST & MLP & 86 \\
    & CNN & 88 \\
    & CapsNet & 89 \\ \midrule
    CIFAR10 & VGG16 & 80 \\
    & NiN & 84 \\
    & ResNet50 & 80 \\ \midrule
    ImageNet & ResNet50 & 76 \\
    \bottomrule
  \end{tabular}
  \label{tab:accuracy}
\end{table}

\subsection{SFIs generated by DE}
\label{sec:resultDE}

\begin{figure}[t]
\centering
\includegraphics[width=\linewidth]{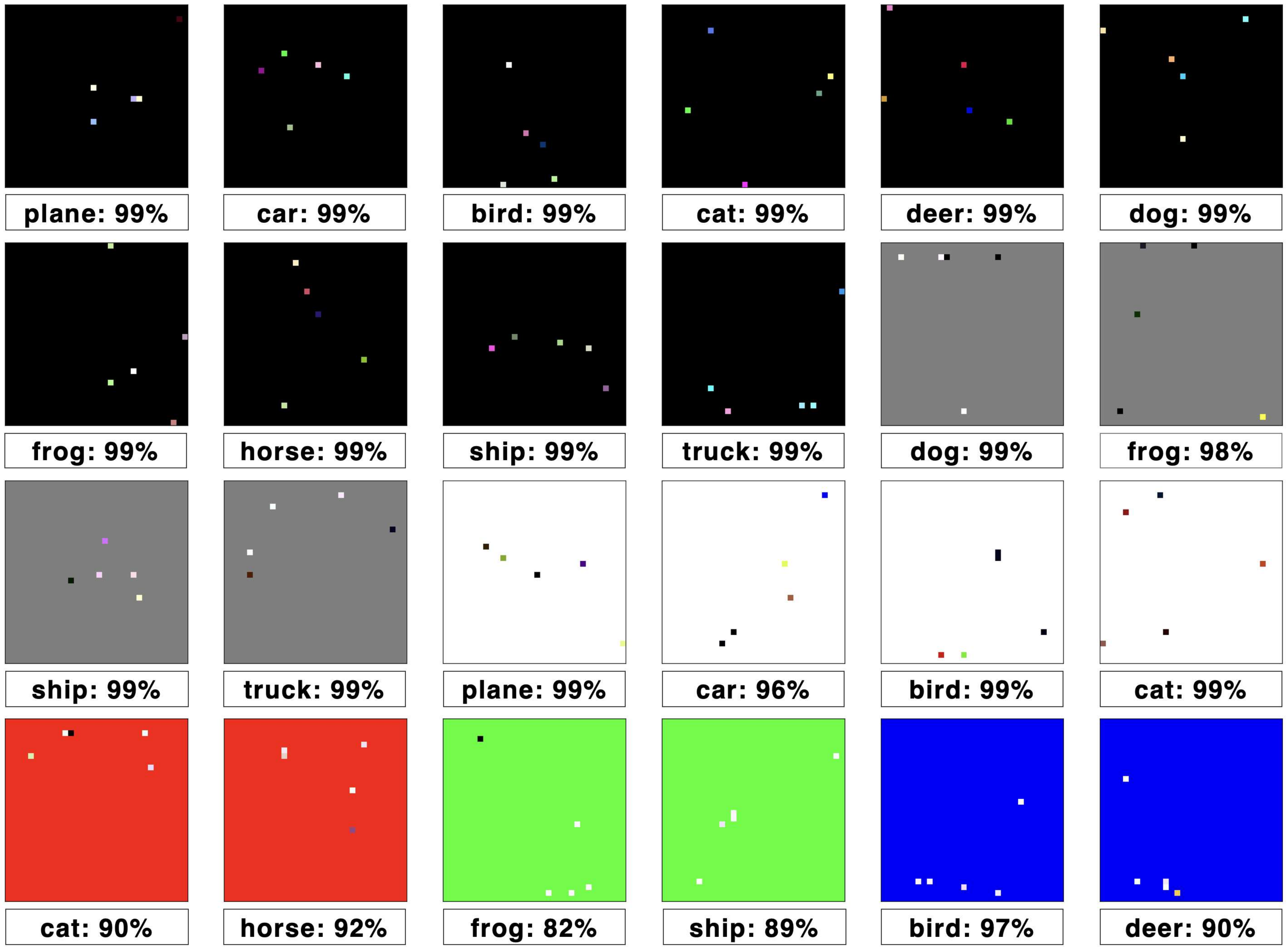}
   \caption{5-SFIs generated by DE. Class name and confidence scores at the bottom of the SFIs are obtained through VGG16~\cite{simonyan2014very} trained on CIFAR10.}
\label{fig:SFI}
\end{figure}

\begin{figure}[t]
\centering
\includegraphics[width=\linewidth]{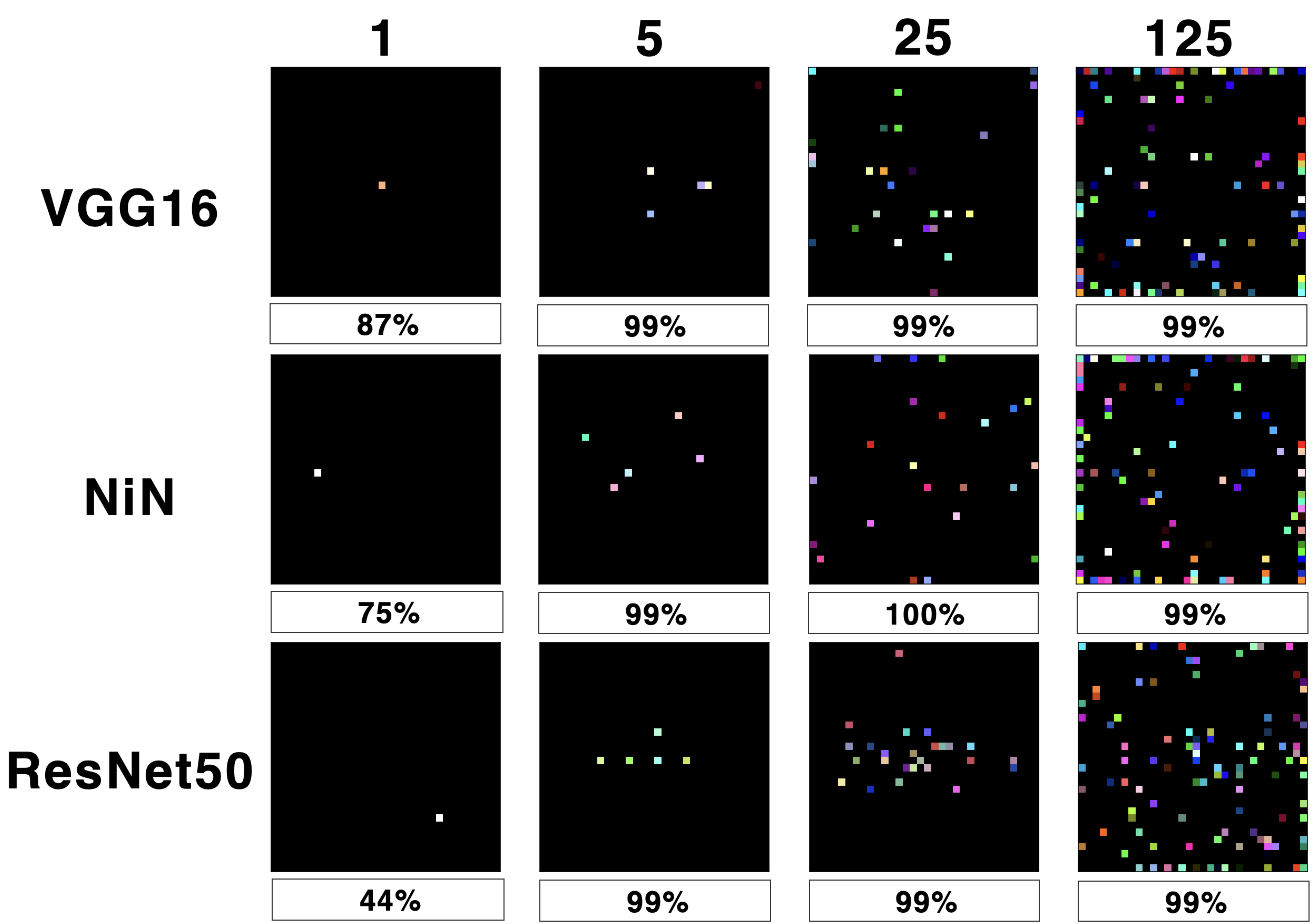}
   \caption{SFIs generated by DE and targeting the plane class. The confidence scores at the bottom of the SFIs are obtained through each model trained on CIFAR10.}
\label{fig:pixelSFI}
\end{figure}

\paragraph{CIFAR10.}
We show that our DE-based method~(cf.~Section~\ref{sec:deMethod}) can generate $k$-SFIs~($k=1,5,25,125$) that can fool the VGG16, NiN, and ResNet50 models trained on CIFAR10. The number of candidate solutions $s$ and iterations of DE $T$ were set to 500 and 3,000, respectively. To evaluate the effect of the base color in SFIs, we performed SFI generation with six base colors~(black, gray, white, red, green, and blue). Figure~\ref{fig:SFI} shows a few examples of SFIs generated by DE. In the top left, \textit{the plane image} classified by the DNN with a high confidence score~(99\%) cannot be recognized by humans as a plane. As shown in Tables~\ref{tab:VGG161}--\ref{tab:VGG16125}, the SFIs generated by the proposed method lead to high confidence scores through the three models over various target classes and base colors.

The confidence scores of a single-color image are not uniform over classes, contrary to intuition. Interestingly, the few pixels that are altered in the base color to generate an SFI have a significant impact on the confidence score through the model. For example, Table~\ref{tab:VGG165} shows a black image is given a 0\% confidence score by the model on the plane class; however, after only five pixel modifications~(SFI at the top left of Figure~\ref{fig:SFI}), the confidence score of the image on the plane class increases to 99\%. The extreme increase in confidence scores by a mere five-pixel change is seen for various base colors and target classes~(\eg, gray and plane class, and white and cat class).

Confidence scores tend to be higher for larger $k$. Figure~\ref{fig:pixelSFI} shows a few examples of SFIs targeting the plane class. Although changing many pixels can produce class-specific features, in this experiment, 25-SFIs and 125-SFIs do not have remarkable patterns, and both yield high confidence scores.

However, some base colors cannot attack the models well regardless of $k$. For example, the confidence scores of blue-based $k$-SFIs~($k=1,5,25,125$) generated for VGG16 and targeting the car class are 0\%~(Tables~\ref{tab:VGG161}--\ref{tab:VGG16125}). We consider the reason why even 125-SFIs cannot fool the models well is because of the incomplete optimization of DE. Our DE-based method takes only a simple calculation, thus it cannot explore the large solution space well. Therefore, it might be hard for the DE-based method to optimize a lot of $k$ pixels to be altered. The difference between the base colors that can attack the model and those that cannot is discussed in Section~\ref{sec:pca}.

\begin{figure}[t]
\centering
\includegraphics[width=\linewidth]{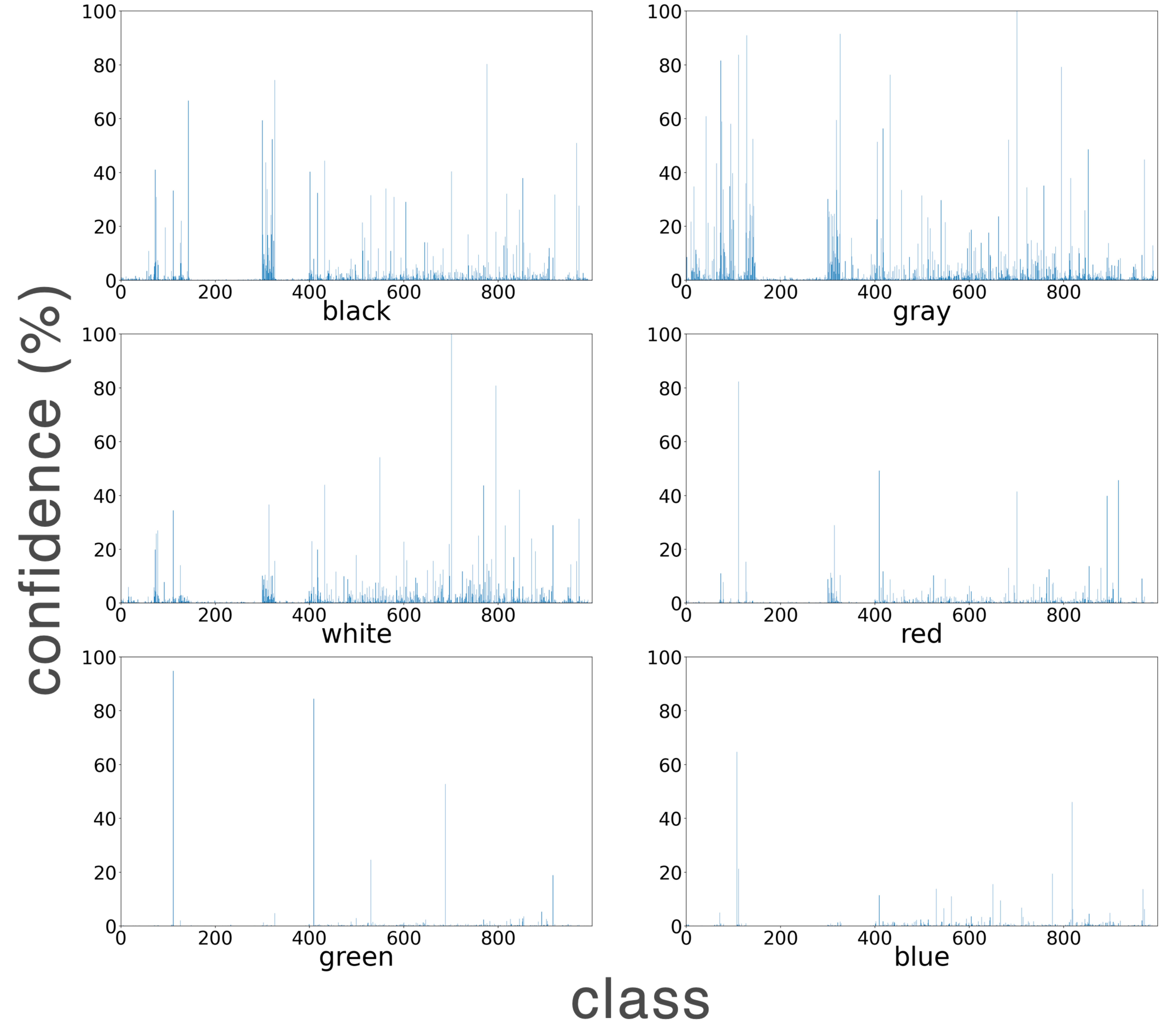}
   \caption{Confidence scores of 5-SFIs generated by DE through ResNet50~\cite{he2016deep} trained on ImageNet. The horizontal axis represents the target class.}
\label{fig:bar}
\end{figure}

\paragraph{ImageNet.}
We also generate 5-SFIs by DE~($s$=500, $T$=500) for ResNet50 trained on ImageNet. The number of iterations of DE, $T$, is set to a lower value than that of the CIFAR10 experiment and the confidence scores were obtained by one SFI because of high computational cost. Figure~\ref{fig:bar} shows almost all SFIs cannot attack the ResNet50 model. In particular, SFIs targeting the class 151--287, dog and cat classes, tend to fail in fooling the model. This might be because of the softmax function. For example, when SFIs target the golden retriever class and try to increase the confidence score of this class, SFIs cannot help increasing the confidence scores of other dog classes such as Brittany spaniel, collie, and dalmatian. Then, outputs corresponding to all dog classes increase, which prohibits the softmax function from giving a high confidence score only to the golden retriever class.

With more iterations of DE~(large $T$), one might be able to yield stronger SFIs that succeed even on ImageNet. We found the gray-based 10-SFI classified as a centipede with a 91\% confidence score through ResNet50 by DE~($s=1,000, T=30,000$)~(Figure~\ref{fig:abst}). Also, we note that we used only 5 pixels for SFIs due to the computational cost. Since 5 pixels is approximately 0.01\%~($5/(224\times224)$) on the whole image, SFIs generation on ImageNet is much more difficult than CIFAR10~(5 pixels is approximately 0.5\% on the whole image). This prevents SFIs from attacking on ImageNet well. We consider that more altered pixels make DNNs more vulnerable on even ImageNet.

\clearpage

\begin{table*}[t]
  \caption{Average confidence scores~(\%) of 1-SFIs through VGG16. Rows denote base colors, and columns denote target classes. That is, the first row and column are the average confidence scores of 50 SFIs targeted to the plane class and are based on a black image. The values in parentheses are the confidence scores of single-color images, such as all black images. The bold text represents values greater than or equal to 90.}
  \centering
  \begin{tabular}{@{}Wc{11mm}Wc{11mm}Wc{11mm}Wc{11mm}Wc{11mm}Wc{11mm}Wc{11mm}Wc{11mm}Wc{11mm}Wc{11mm}Wc{11mm}@{}}
    \toprule
    & plane & car & bird & cat & deer & dog & frog & horse & ship & truck \\ \midrule
    black & 87~(0) & 7~(0) & \textbf{99}~(0) & \textbf{99}~(62) & \textbf{98}~(0) & 37~(0) & \textbf{98}~(35) & 79~(0) & 59~(0) & \textbf{94}~(0) \\
    gray & \textbf{99}~(11) & 0~(0) & \textbf{92}~(2) & 63~(3) & 7~(0) & 0~(0) & 17~(0) & 0~(0) & \textbf{99}~(81) & 5~(0) \\
    white & \textbf{99}~(9) & 1~(0) & \textbf{99}~(5) & 87~(0) & 11~(0) & 0~(0) & 17~(0) & 9~(0) & \textbf{99}~(84) & 44~(0) \\
    red & \textbf{99}~(\textbf{97}) & 0~(0) & 0~(0) & 16~(0) & 0~(0) & 0~(0) & 0~(0) & 1~(0) & 6~(0) & 8~(0) \\
    green & 18~(2) & 6~(3) & 2~(0) & 70~(50) & 1~(0) & 0~(0) & 12~(3) & 0~(0) & 48~(20) & 32~(17) \\
    blue & \textbf{99}~(\textbf{95}) & 0~(0) & 22~(2) & 0~(0) & 0~(0) & 0~(0) & 0~(0) & 0~(0) & 24~(1) & 0~(0) \\
    \bottomrule
  \end{tabular}
  \label{tab:VGG161}
\end{table*}

\begin{table*}[t]
  \caption{Average confidence scores~(\%) of 5-SFIs through VGG16. The description of the table are the same as Table~\ref{tab:VGG161}.}
  \centering
  \small
  \begin{tabular}{@{}Wc{11mm}Wc{11mm}Wc{11mm}Wc{11mm}Wc{11mm}Wc{11mm}Wc{11mm}Wc{11mm}Wc{11mm}Wc{11mm}Wc{11mm}@{}}
    \toprule
    & plane & car & bird & cat & deer & dog & frog & horse & ship & truck \\ \midrule
    black & \textbf{99}~(0) & \textbf{99}~(0) & \textbf{99}~(0) & \textbf{99}~(62) & \textbf{99}~(0) & \textbf{99}~(0) & \textbf{99}~(35) & \textbf{99}~(0) & \textbf{99}~(0) & \textbf{99}~(0) \\
    gray  & \textbf{100}~(11) & 52~(0) & \textbf{99}~(2) & \textbf{99}~(3) & \textbf{99}~(0)& \textbf{95}~(0) & \textbf{99}~(0) & \textbf{99}~(0) & \textbf{99}~(81) & \textbf{97}~(0) \\
    white & \textbf{100}~(9) & 33~(0) & \textbf{100}~(5) & \textbf{99}~(0) & \textbf{99}~(0)& \textbf{99}~(0) & \textbf{98}~(0) & \textbf{99}~(0) & \textbf{99}~(84) & \textbf{99}~(0) \\
    red   & \textbf{99}~(\textbf{97}) & 53~(0) & 56~(0) & \textbf{98}~(0) & 4~(0)& 0~(0) & 1~(0) & 17~(0) & \textbf{97}~(0) & \textbf{91}~(0) \\
    green & \textbf{99}~(2) & 29~(3) & 74~(0) & \textbf{96}~(50) & 52~(0)& 3~(0) & \textbf{90}~(3) & 5~(0) & \textbf{92}~(20) & 87~(17) \\
    blue  & \textbf{99}~(\textbf{95}) & 0~(0) & \textbf{98}~(2) & 37~(0) & 4~(0)& 0~(0) & 0~(0) & 0~(0) & \textbf{98}~(1) & 0~(0) \\
    \bottomrule
  \end{tabular}
  \label{tab:VGG165} 
\end{table*}

\begin{table*}[t]
  \caption{Average confidence scores~(\%) of 25-SFIs through VGG16. The description of the table are the same as Table~\ref{tab:VGG161}.}
  \centering
  \begin{tabular}{@{}Wc{11mm}Wc{11mm}Wc{11mm}Wc{11mm}Wc{11mm}Wc{11mm}Wc{11mm}Wc{11mm}Wc{11mm}Wc{11mm}Wc{11mm}@{}}
    \toprule
    & plane & car & bird & cat & deer & dog & frog & horse & ship & truck \\ \midrule
    black & \textbf{99}~(0) & \textbf{99}~(0) & \textbf{99}~(0) & \textbf{99}~(62) & \textbf{99}~(0) & \textbf{99}~(0) & \textbf{99}~(35) & \textbf{99}~(0) & \textbf{99}~(0) & \textbf{99}~(0) \\
    gray & \textbf{100}~(11) & 66~(0) & \textbf{99}~(2) & \textbf{99}~(3) & \textbf{99}~(0) & \textbf{97}~(0) & \textbf{99}~(0) & \textbf{97}~(0) & \textbf{99}~(81) & \textbf{98}~(0) \\
    white & \textbf{100}~(9) & \textbf{90}~(0) & \textbf{99}~(5) & \textbf{99}~(0) & \textbf{99}~(0) & \textbf{99}~(0) & \textbf{98}~(0) & \textbf{99}~(0) & \textbf{99}~(84) & \textbf{99}~(0) \\
    red & \textbf{99}~(\textbf{97}) & 77~(0) & 52~(0) & \textbf{99}~(0) & 2~(0) & 1~(0) & 4~(0) & 32~(0) & 89~(0) & 85~(0) \\
    green & \textbf{96}~(2) & 26~(3) & 31~(0) & \textbf{97}~(50) & 21~(0) & 0~(0) & \textbf{96}~(3) & 4~(0) & 84~(20) & 74~(17) \\
    blue & \textbf{99}~(\textbf{95}) & 0~(0) & \textbf{99}~(2) & 66~(0) & 5~(0) & 0~(0) & 1~(0) & 0~(0) & \textbf{99}~(1) & 0~(0) \\
    \bottomrule
  \end{tabular}
  \label{tab:VGG1625}
\end{table*}

\begin{table*}[t]
  \caption{Average confidence scores~(\%) of 125-SFIs through VGG16. The description of the table are the same as Table~\ref{tab:VGG161}.}
  \centering
  \begin{tabular}{@{}Wc{11mm}Wc{11mm}Wc{11mm}Wc{11mm}Wc{11mm}Wc{11mm}Wc{11mm}Wc{11mm}Wc{11mm}Wc{11mm}Wc{11mm}@{}}
    \toprule
    & plane & car & bird & cat & deer & dog & frog & horse & ship & truck \\ \midrule
    black & \textbf{99}~(0) & \textbf{99}~(0) & \textbf{99}~(0) & \textbf{99}~(62) & \textbf{99}~(0) & \textbf{99}~(0) & \textbf{99}~(35) & \textbf{99}~(0) & \textbf{99}~(0) & \textbf{99}~(0) \\
    gray & \textbf{99}~(11) & 81~(0) & \textbf{99}~(2) & \textbf{99}~(3) & \textbf{99}~(0) & \textbf{99}~(0) & \textbf{99}~(0) & \textbf{99}~(0) & \textbf{99}~(81) & \textbf{99}~(0) \\
    white & \textbf{99}~(9) & \textbf{98}~(0) & \textbf{99}~(5) & \textbf{99}~(0) & \textbf{99}~(0) & \textbf{99}~(0) & \textbf{99}~(0) & \textbf{99}~(0) & \textbf{99}~(84) & \textbf{99}~(0) \\
    red & \textbf{99}~(\textbf{97}) & \textbf{98}~(0) & \textbf{97}~(0) & \textbf{99}~(0) & 77~(0) & 25~(0) & \textbf{92}~(0) & 79~(0) & \textbf{94}~(0) & \textbf{99}~(0) \\
    green & \textbf{98}~(2) & 30~(3) & 65~(0) & \textbf{98}~(50) & 73~(0) & 0~(0) & \textbf{99}~(3) & 12~(0) & 85~(20) & 87~(17) \\
    blue & \textbf{99}~(\textbf{95}) & 0~(0) & \textbf{99}~(2) & \textbf{90}~(0) & 7~(0) & 0~(0) & 3~(0) & 0~(0) & \textbf{99}~(1) & 0~(0) \\
    \bottomrule
  \end{tabular}
  \label{tab:VGG16125}
\end{table*}

\clearpage

\begin{table*}[t]
  \caption{Average confidence scores~(\%) of 1-SFIs through NiN~\cite{lin2013network}. The description of the table are the same as Table~\ref{tab:VGG161}.}
  \centering
  \begin{tabular}{@{}Wc{11mm}Wc{11mm}Wc{11mm}Wc{11mm}Wc{11mm}Wc{11mm}Wc{11mm}Wc{11mm}Wc{11mm}Wc{11mm}Wc{11mm}@{}}
    \toprule
    & plane & car & bird & cat & deer & dog & frog & horse & ship & truck \\ \midrule
    black & 75~(7) & 40~(1) & 62~(25) & 83~(33) & 38~(22) & 4~(2) & 3~(2) & 5~(0) & 8~(3) & 3~(0) \\
    gray & \textbf{93}~(35) & 11~(1) & 56~(12) & 11~(3) & 10~(5) & 2~(0) & 3~(2) & 0~(0) & 42~(36) & 3~(0) \\
    white & \textbf{96}~(27) & 2~(0) & 85~(35) & 23~(6) & 25~(13) & 1~(0) & 4~(2) & 1~(0) & 26~(12) & 0~(0) \\
    red & 67~(5) & 68~(26) & 6~(1) & 44~(28) & 20~(11) & 14~(9) & 2~(1) & 17~(7) & 2~(1) & 27~(6) \\
    green & 63~(11) & 2~(1) & 82~(23) & 7~(3) & 53~(30) & 0~(0) & 38~(19) & 0~(0) & 18~(9) & 0~(0) \\
    blue & \textbf{95}~(20) & 4~(2) & 9~(2) & 13~(5) & 23~(14) & 1~(0) & 6~(2) & 0~(0) & 63~(50) & 1~(1) \\
    \bottomrule
  \end{tabular}
  \label{tab:NiN1}
\end{table*}

\begin{table*}[t]
  \caption{Average confidence scores~(\%) of 5-SFIs through NiN. The description of the table are the same as Table~\ref{tab:VGG161}.}
  \centering
  \begin{tabular}{@{}Wc{11mm}Wc{11mm}Wc{11mm}Wc{11mm}Wc{11mm}Wc{11mm}Wc{11mm}Wc{11mm}Wc{11mm}Wc{11mm}Wc{11mm}@{}}
    \toprule
    & plane & car & bird & cat & deer & dog & frog & horse & ship & truck \\ \midrule
    black & \textbf{99}~(7) & \textbf{98}~(1) & \textbf{99}~(25) & \textbf{99}~(33) & 81~(22) & 86~(2) & 17~(2) & 82~(0) & \textbf{90}~(3) & 81~(0) \\
    gray & \textbf{99}~(35) & 85~(1) & \textbf{99}~(12) & 69~(3) & 41~(5) & \textbf{98}~(0) & 20~(2) & 17~(0) & \textbf{97}~(36) & 44~(0) \\
    white & \textbf{100}~(27) & 36~(0) & \textbf{99}~(35) & \textbf{95}~(6) & 72~(13) & \textbf{99}~(0) & 18~(2) & 23~(0) & \textbf{98}~(12) & 4~(0) \\
    red & \textbf{99}~(5) & \textbf{97}~(26) & 74~(1) & 88~(28) & 68~(11) & 52~(9) & 15~(1) & \textbf{97}~(7) & 17~(1) & \textbf{93}~(6) \\
    green & \textbf{99}~(11) & 16~(1) & \textbf{99}~(23) & 24~(3) & 88~(30) & 0~(0) & 84~(19) & 2~(0) & 53~(9) & 4~(0) \\
    blue & \textbf{99}~(20) & 27~(2) & 64~(2) & 83~(5) & 60~(14) & 12~(0) & 40~(2) & 5~(0) & \textbf{98}~(50) & 27~(1) \\
    \bottomrule
  \end{tabular}
  \label{tab:NiN5}
\end{table*}

\begin{table*}[t]
  \caption{Average confidence scores~(\%) of 25-SFIs through NiN. The description of the table are the same as Table~\ref{tab:VGG161}.}
  \centering
  \begin{tabular}{@{}Wc{11mm}Wc{11mm}Wc{11mm}Wc{11mm}Wc{11mm}Wc{11mm}Wc{11mm}Wc{11mm}Wc{11mm}Wc{11mm}Wc{11mm}@{}}
    \toprule
    & plane & car & bird & cat & deer & dog & frog & horse & ship & truck \\ \midrule
    black & \textbf{100}~(7) & \textbf{99}~(1) & \textbf{99}~(25) & \textbf{99}~(33) & \textbf{95}~(22) & \textbf{98}~(2) & \textbf{98}~(2) & \textbf{95}~(0) & \textbf{99}~(3) & \textbf{99}~(0) \\
    gray & \textbf{100}~(35) & \textbf{97}~(1) & \textbf{99}~(12) & \textbf{95}~(3) & 86~(5) & \textbf{98}~(0) & 87~(2) & 59~(0) & \textbf{98}~(36) & 77~(0) \\
    white & \textbf{100}~(27) & \textbf{97}~(0) & \textbf{99}~(35) & \textbf{99}~(6) & \textbf{90}~(13) & \textbf{99}~(0) & 51~(2) & \textbf{98}~(0) & \textbf{99}~(12) & 58~(0) \\
    red & \textbf{99}~(5) & \textbf{99}~(26) & 68~(1) & \textbf{93}~(28) & 81~(11) & 63~(9) & 22~(1) & 70~(7) & 26~(1) & \textbf{98}~(6) \\
    green & \textbf{99}~(11) & 27~(1) & \textbf{99}~(23) & 39~(3) & 88~(30) & 0~(0) & \textbf{99}~(19) & 2~(0) & 29~(9) & 2~(0) \\
    blue & \textbf{99}~(20) & 55~(2) & 83~(2) & \textbf{96}~(5) & 77~(14) & 24~(0) & 39~(2) & 18~(0) & \textbf{97}~(50) & 42~(1) \\
    \bottomrule
  \end{tabular}
  \label{tab:NiN25}
\end{table*}

\begin{table*}[t]
  \caption{Average confidence scores~(\%) of 125-SFIs through NiN. The description of the table are the same as Table~\ref{tab:VGG161}.}
  \centering
  \begin{tabular}{@{}Wc{11mm}Wc{11mm}Wc{11mm}Wc{11mm}Wc{11mm}Wc{11mm}Wc{11mm}Wc{11mm}Wc{11mm}Wc{11mm}Wc{11mm}@{}}
    \toprule
    & plane & car & bird & cat & deer & dog & frog & horse & ship & truck \\ \midrule
    black & \textbf{99}~(7) & \textbf{99}~(1) & \textbf{99}~(25) & \textbf{99}~(33) & \textbf{99}~(22) & \textbf{95}~(2) & \textbf{99}~(2) & \textbf{93}~(0) & \textbf{99}~(3) & \textbf{99}~(0) \\
    gray & \textbf{99}~(35) & \textbf{99}~(1) & \textbf{99}~(12) & \textbf{99}~(3) & \textbf{99}~(5) & \textbf{98}~(0) & \textbf{99}~(2) & 82~(0) & \textbf{99}~(36) & \textbf{96}~(0) \\
    white & \textbf{100}~(27) & \textbf{99}~(0) & \textbf{99}~(35) & \textbf{99}~(6) & \textbf{99}~(13) & \textbf{99}~(0) & \textbf{99}~(2) & \textbf{96}~(0) & \textbf{99}~(12) & \textbf{99}~(0) \\
    red & \textbf{99}~(5) & \textbf{99}~(26) & \textbf{96}~(1) & \textbf{99}~(28) & 89~(11) & 77~(9) & 86~(1) & \textbf{94}~(7) & 64~(1) & \textbf{99}~(6) \\
    green & \textbf{98}~(11) & 47~(1) & \textbf{99}~(23) & 74~(3) & \textbf{99}~(30) & 12~(0) & \textbf{99}~(19) & 8~(0) & 24~(9) & 2~(0) \\
    blue & \textbf{99}~(20) & \textbf{99}~(2) & \textbf{98}~(2) & \textbf{99}~(5) & 88~(14) & 55~(0) & 70~(2) & 41~(0) & \textbf{99}~(50) & \textbf{99}~(1) \\
    \bottomrule
  \end{tabular}
  \label{tab:NiN125}
\end{table*}

\clearpage

\begin{table*}[t]
  \caption{Average confidence scores~(\%) of 1-SFIs through ResNet50. The description of the table are the same as Table~\ref{tab:VGG161}.}
  \centering
  \begin{tabular}{@{}Wc{11mm}Wc{11mm}Wc{11mm}Wc{11mm}Wc{11mm}Wc{11mm}Wc{11mm}Wc{11mm}Wc{11mm}Wc{11mm}Wc{11mm}@{}}
    \toprule
    & plane & car & bird & cat & deer & dog & frog & horse & ship & truck \\ \midrule
    black & 23~(8) & 0~(0) & 16~(5) & 56~(13) & 83~(65) & 0~(0) & 1~(0) & 2~(0) & 12~(5) & 0~(0) \\
    gray & \textbf{99}~(11) & 0~(0) & 16~(1) & 5~(0) & 16~(3) & 4~(0) & 2~(0) & 0~(0) & \textbf{96}~(81) & 0~(0) \\
    white & \textbf{90}~(71) & 0~(0) & 22~(13) & 23~(11) & 0~(0) & 0~(0) & 0~(0) & 0~(0) & 4~(2) & 0~(0) \\
    red & 11~(8) & 18~(14) & 5~(3) & 67~(62) & 0~(0) & 1~(0) & 0~(0) & 0~(0) & 0~(0) & 10~(8) \\
    green & \textbf{99}~(\textbf{99}) & 0~(0) & 0~(0) & 0~(0) & 0~(0) & 0~(0) & 0~(0) & 0~(0) & 0~(0) & 0~(0) \\
    blue & \textbf{90}~(65) & 0~(0) & 34~(15) & 3~(1) & 11~(6) & 0~(0) & 2~(0) & 0~(0) & 17~(9) & 0~(0) \\
  \end{tabular}
  \label{tab:ResNet501}
\end{table*}

\begin{table*}[t]
  \caption{Average confidence scores~(\%) of 5-SFIs through ResNet50. The description of the table are the same as Table~\ref{tab:VGG161}.}
  \centering
  \begin{tabular}{@{}Wc{11mm}Wc{11mm}Wc{11mm}Wc{11mm}Wc{11mm}Wc{11mm}Wc{11mm}Wc{11mm}Wc{11mm}Wc{11mm}Wc{11mm}@{}}
    \toprule
    & plane & car & bird & cat & deer & dog & frog & horse & ship & truck \\ \midrule
    black & \textbf{99}~(8) & 0~(0) & 88~(5) & \textbf{97}~(13) & \textbf{96}~(65) & 0~(0) & 11~(0) & 67~(0) & 34~(5) & 8~(0) \\
    gray & \textbf{99}~(11) & 2~(0) & \textbf{99}~(1) & 81~(0) & 85~(3) & \textbf{97}~(0) & 86~(0) & 0~(0) & \textbf{99}~(81) & 34~(0) \\
    white & \textbf{99}~(71) & 1~(0) & \textbf{99}~(13) & \textbf{90}~(11) & 10~(0) & 6~(0) & 41~(0) & 0~(0) & 70~(2) & 7~(0) \\
    red & 18~(8) & 70~(14) & 9~(3) & 72~(62) & 0~(0) & 1~(0) & 0~(0) & 0~(0) & 0~(0) & 14~(8) \\
    green & \textbf{99}~(\textbf{99}) & 0~(0) & 0~(0) & 0~(0) & 0~(0) & 0~(0) & 4~(0) & 0~(0) & 0~(0) & 0~(0) \\
    blue & \textbf{99}~(65) & 0~(0) & \textbf{94}~(15) & 15~(1) & 18~(6) & 0~(0) & 9~(0) & 0~(0) & 53~(9) & 0~(0) \\
    \bottomrule
  \end{tabular}
  \label{tab:ResNet505}
\end{table*}

\begin{table*}[t]
  \caption{Average confidence scores~(\%) of 25-SFIs through ResNet50. The description of the table are the same as Table~\ref{tab:VGG161}.}
  \centering
  \begin{tabular}{@{}Wc{11mm}Wc{11mm}Wc{11mm}Wc{11mm}Wc{11mm}Wc{11mm}Wc{11mm}Wc{11mm}Wc{11mm}Wc{11mm}Wc{11mm}@{}}
    \toprule
    & plane & car & bird & cat & deer & dog & frog & horse & ship & truck \\ \midrule
    black & \textbf{99}~(8) & 33~(0) & \textbf{99}~(5) & \textbf{99}~(13) & \textbf{99}~(65) & 4~(0) & 84~(0) & \textbf{99}~(0) & 77~(5) & \textbf{93}~(0) \\
    gray & \textbf{99}~(11) & 38~(0) & \textbf{99}~(1) & \textbf{98}~(0) & \textbf{99}~(3) & \textbf{99}~(0) & \textbf{99}~(0) & 6~(0) & \textbf{99}~(81) & 36~(0) \\
    white & \textbf{99}~(71) & 6~(0) & \textbf{99}~(13) & \textbf{99}~(11) & \textbf{96}~(0) & 49~(0) & \textbf{99}~(0) & 9~(0) & 89~(2) & 19~(0) \\
    red & 27~(8) & 84~(14) & 11~(3) & 77~(62) & 0~(0) & 1~(0) & 0~(0) & 0~(0) & 1~(0) & 18~(8) \\
    green & \textbf{99}~(\textbf{99}) & 0~(0) & 0~(0) & 0~(0) & 0~(0) & 0~(0) & 18~(0) & 0~(0) & 0~(0) & 0~(0) \\
    blue & \textbf{99}~(65) & 0~(0) & \textbf{96}~(15) & 49~(1) & 33~(6) & 1~(0) & 26~(0) & 1~(0) & 63~(9) & 0~(0) \\
    \bottomrule
  \end{tabular}
  \label{tab:ResNet5025}
\end{table*}

\begin{table*}[t]
  \caption{Average confidence scores~(\%) of 125-SFIs through ResNet50. The description of the table are the same as Table~\ref{tab:VGG161}.}
  \centering
  \begin{tabular}{@{}Wc{11mm}Wc{11mm}Wc{11mm}Wc{11mm}Wc{11mm}Wc{11mm}Wc{11mm}Wc{11mm}Wc{11mm}Wc{11mm}Wc{11mm}@{}}
    \toprule
    & plane & car & bird & cat & deer & dog & frog & horse & ship & truck \\ \midrule
    black & \textbf{99}~(8) & \textbf{99}~(0) & \textbf{99}~(5) & \textbf{99}~(13) & \textbf{99}~(65) & \textbf{99}~(0) & \textbf{99}~(0) & \textbf{98}~(0) & \textbf{95}~(5) & \textbf{99}~(0) \\
    gray & \textbf{99}~(11) & \textbf{98}~(0) & \textbf{99}~(1) & \textbf{99}~(0) & \textbf{99}~(3) & \textbf{99}~(0) & \textbf{99}~(0) & 89~(0) & \textbf{99}~(81) & 87~(0) \\
    white & \textbf{99}~(71) & 59~(0) & \textbf{99}~(13) & \textbf{99}~(11) & \textbf{99}~(0) & \textbf{99}~(0) & \textbf{100}~(0) & \textbf{98}~(0) & \textbf{94}~(2) & 32~(0) \\
    red & \textbf{99}~(8) & \textbf{92}~(14) & 24~(3) & 82~(62) & 2~(0) & 4~(0) & 2~(0) & 0~(0) & 1~(0) & 65~(8) \\
    green & \textbf{99}~(\textbf{99}) & 44~(0) & 0~(0) & 0~(0) & 1~(0) & 0~(0) & \textbf{97}~(0) & 0~(0) & 0~(0) & 0~(0) \\
    blue & \textbf{100}~(65) & 9~(0) & \textbf{99}~(15) & \textbf{96}~(1) & 81~(6) & 25~(0) & \textbf{96}~(0) & 43~(0) & 80~(9) & 12~(0) \\
    \bottomrule
  \end{tabular}
  \label{tab:ResNet50125}
\end{table*}

\clearpage

\begin{figure}[t]
\centering
\includegraphics[width=\linewidth]{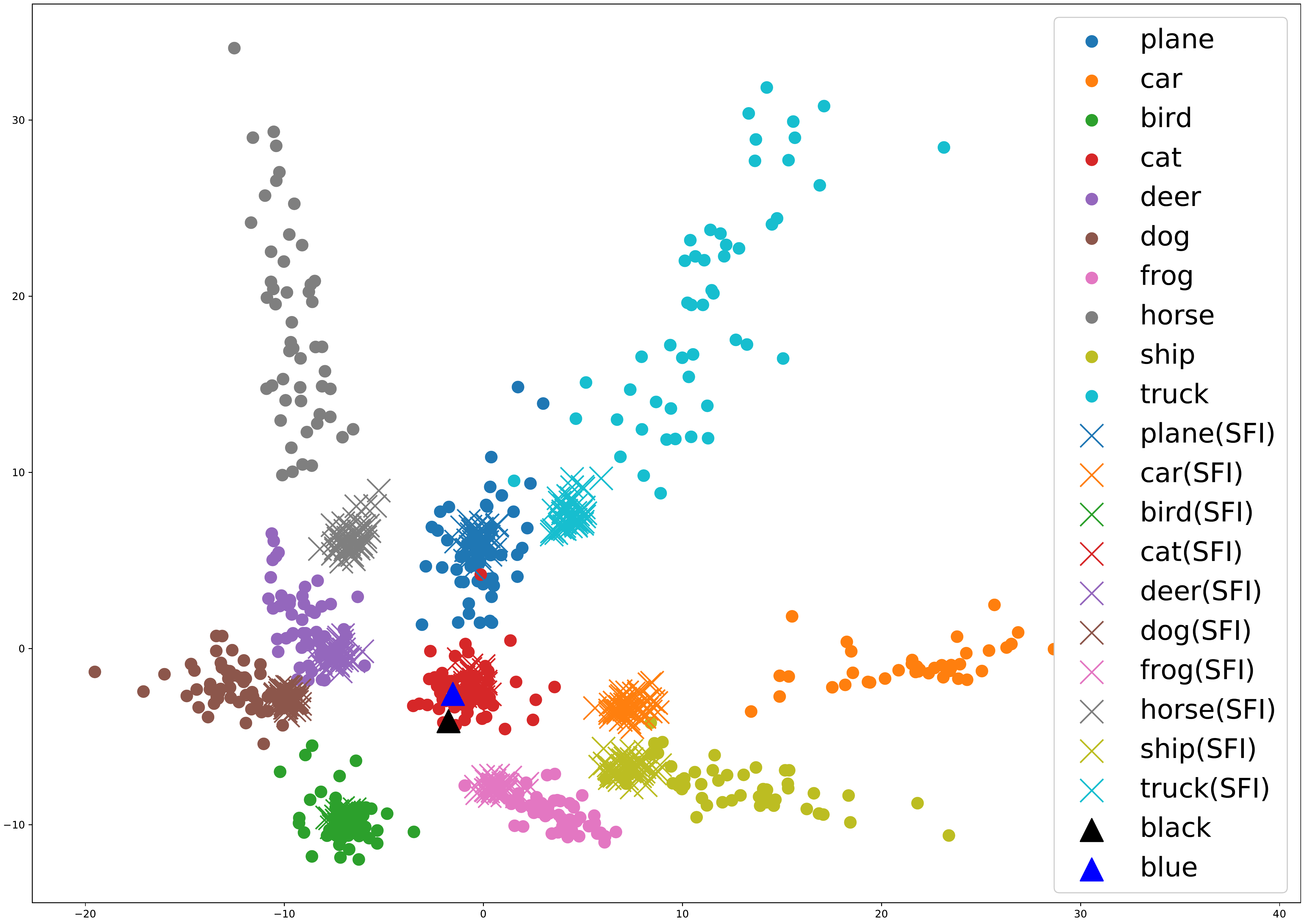}
   \caption{Result of PCA on a total of 1,002 images~(50 natural images per class, 50 black-based SFIs generated by DE per class, a black image, and a blue image). Deep features of natural images and SFIs through VGG16 are close to each other in the deep layer.}
\label{fig:PCA}
\end{figure}

\begin{figure}[t]
\centering
\includegraphics[width=\linewidth]{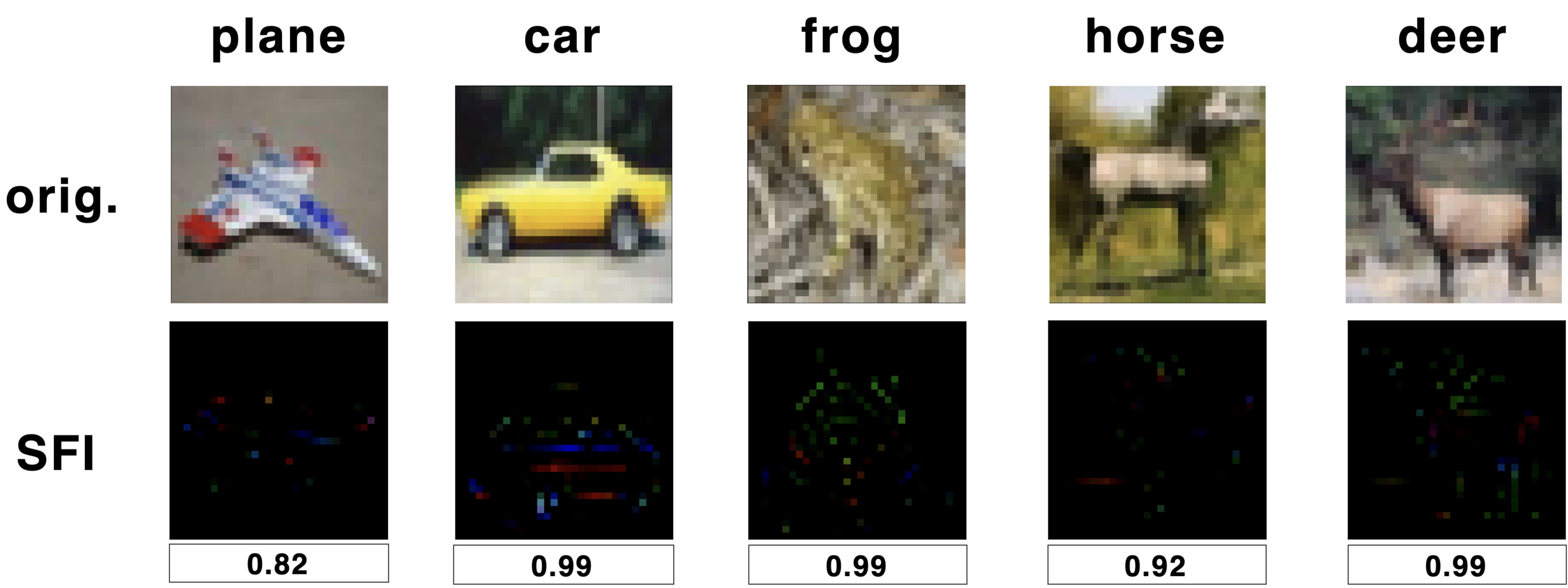}
   \caption{Result of inverting a natural image~(orig.) into an SFI. Even though these images are extremely different in appearance, they have similar features in the deep layers of the model. The values below the image represent the inner products between the original image and the SFI.}
\label{fig:invert}
\end{figure}

\subsection{Features of SFIs in the deep layer}
\label{sec:pca}
To understand why changing only a few pixels in a single-color image can significantly alter the confidence scores through a model, we compared the deep features of natural images and those of five black-based SFIs generated by DE. Specifically, we compared the output at the second-to-last layer of the VGG16 model trained on CIFAR10 for natural images and SFIs. Deep features are projected into a two-dimensional space through a principal component analysis~(PCA)~(Figure~\ref{fig:PCA}). It can be seen that within the same class~(\eg, airplane class), both natural images~(blue dots) and SFIs~(blue crosses) are distributed nearby. This result shows that natural images and SFIs, which appear to be completely different in the input space, become extremely similar after passing through the layers of the model. This result also indicates the difference between a base color that can attack~(black triangle) or cannot attack~(blue triangle)~(cf.~Section~\ref{sec:resultDE}) does not depend on the position of the deep features of the base color. The difference will be studied in a future study; however, an improved DE~\cite{brest2006self, vargas2015general} may mitigate the effect of the base color because the difference might be attributed to a defective optimization.

We emphasize that this result is not trivial; that is, in the feature space, SFIs could be distributed far from natural images and classified as the target class with high confidence scores. In fact, some OOD methods take advantage of the differences in the distribution of deep features of the in-distribution objects and OOD objects of the same class~\cite{hendrycks2016baseline, lee2018simple}. The PCA projection results show that natural images and SFIs are distributed near the feature space. Therefore, they cannot be discriminated from each other by such OOD methods.

\subsection{SFIs generated by deep feature inversion}
\label{sec:invert}
Natural images and SFIs that are given the same label by a model are distributed near the feature space. This observation implies that, given a natural image, we can generate an SFI that is extremely similar to the given image in the feature space, but not in the input space. Herein, we demonstrate that our second SFI generation method~(cf. Section~\ref{sec:invertMethod}) realizes this. Generating SFIs, we set $\lambda_{\mathrm{sparse}}$ to $1.0\times10^{-5}$ and $z_b$ to 0~(\ie, black-based SFIs) in~(\ref{eq:natural2pixel}). We used the deep features in the second to last layer of the VGG16 model trained on CIFAR10. A few examples of SFIs generated by deep feature inversion are shown in Figure~\ref{fig:invert}. The corresponding SFI can be generated for a given natural image. Interestingly, some colors of the altered pixels of the SFIs generated by deep feature inversion are not included in the original image. For example, the SFI generated from the yellow car image in the second column from the left in Figure~\ref{fig:invert} has red and blue pixels that do not appear to be included in the original image.

\begin{figure}[t]
\centering
\includegraphics[width=\linewidth]{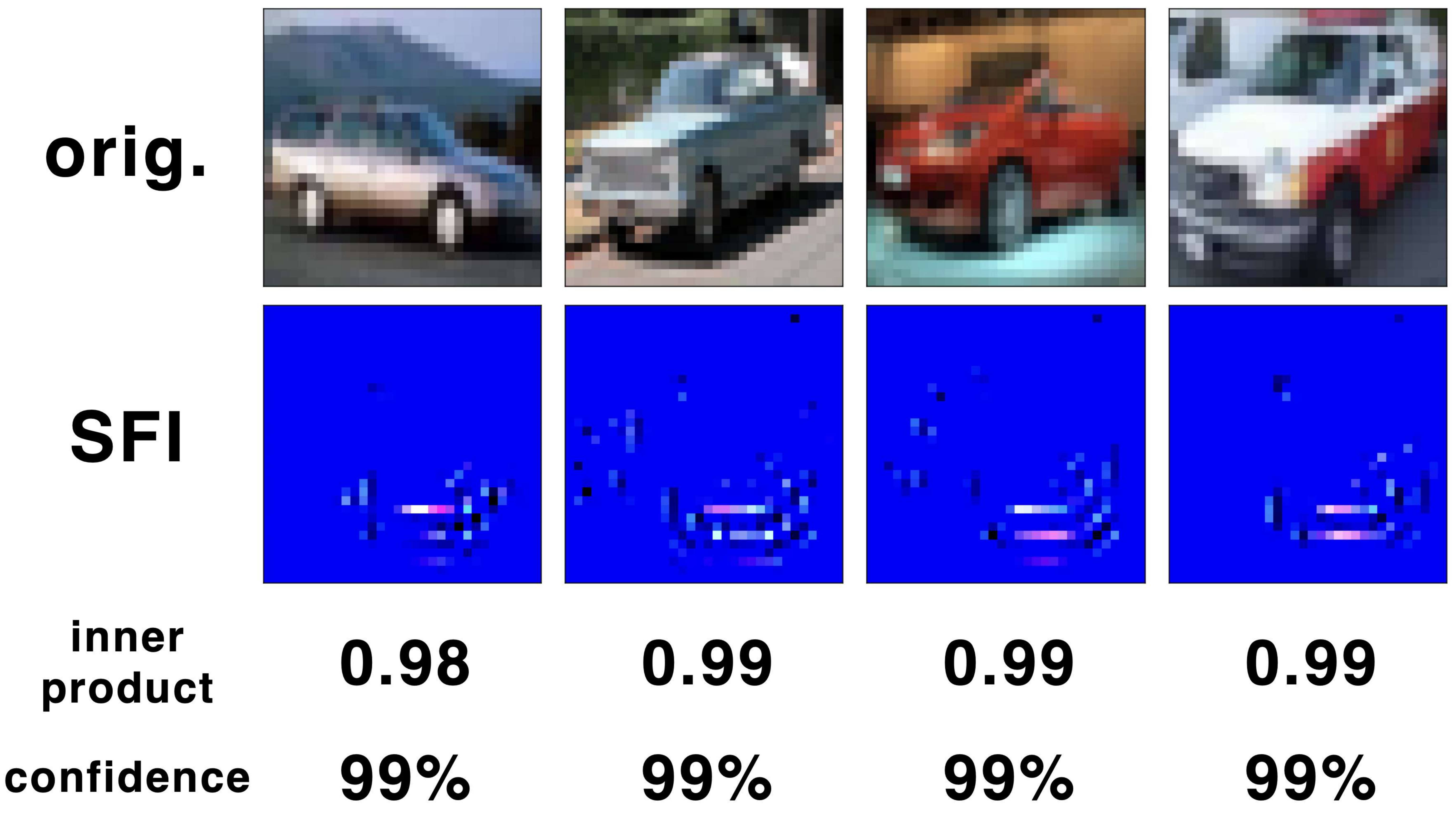}
   \caption{Original natural images and corresponding SFIs generated by deep feature inversion. The inner product is obtained from deep features of an SFI and those of the original natural image. The confidence scores are obtained by SFIs through the VGG16 model.}
\label{fig:blueinversion}
\end{figure}

Blue-based SFIs generated by DE can attack few classes of CIFAR10 through VGG16 model~(Table~\ref{tab:VGG161}--\ref{tab:VGG16125}). To understand the difference between our two methods, we generate blue-based SFIs by deep feature inversion. In particular, we target the car class that SFIs generated by DE cannot fool at all through VGG16. Figure~\ref{fig:blueinversion} shows the original natural car images and corresponding SFIs. Deep feature inversion can generate blue-based SFIs targeting the car class. This is because deep feature inversion in a image-targeted setting can treat many altered pixels more efficiently than DE in a class-targeted setting. It is left for future work whether SFIs with only five altered pixels can fool the DNN or not.

\subsection{Comparison between DE-based and deep feature inversion methods}
\begin{figure*}[t]
\centering
\includegraphics[width=\linewidth]{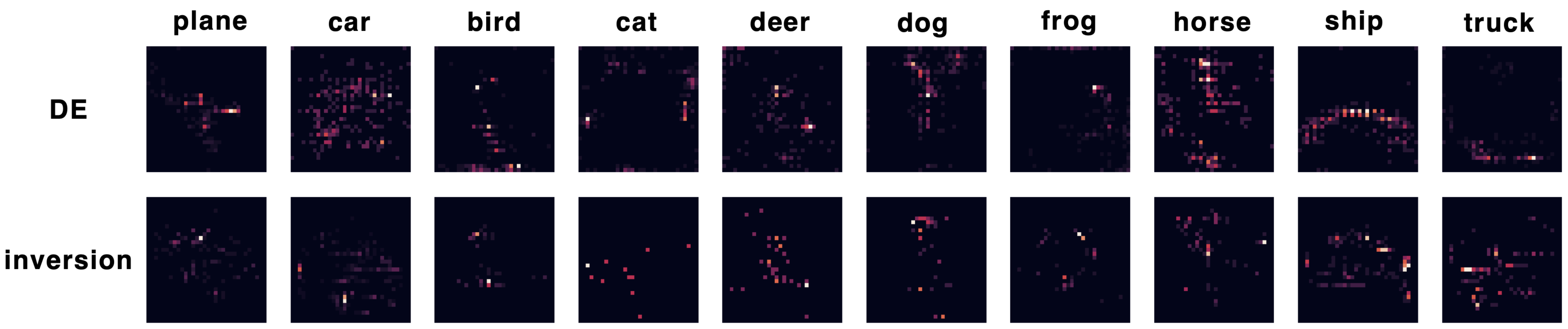}
   \caption{Distribution of changed pixels~(intensity$>$0.3, where the range is $[0, 1]$) in SFIs generated by each method, represented as heat maps. Black, red, and white values indicate increases in that order.}
\label{fig:distribution}
\end{figure*}

The distribution of the modified pixel positions in the SFIs generated by DE and deep feature inversion are compared in Figure~\ref{fig:distribution}. Some classes~(car, bird, deer, dog, horse, and ship) seem to have inclusion relations, and the other classes~(plane, cat, frog, and truck) do not seem to match at all. Some pixel positions are shared. For example, there are white dots~(\ie, positions where modified pixels are likely to be present) on the left in the cat class, at the right in the deer class, and at the center and top in the bird class, in both DE and deep feature inversion. The result in which the prediction of the DNN relies heavily on some pixels suggests the vulnerability of the DNN to parts of an image.

\subsection{Comparison of four architectures}
We now discuss why SFIs can attack DNNs successfully by comparing four different architectures: MLP, CNN$_\text{MP}$, CNN$_\text{None}$, and CapsNet. MLP consists of three fully connected layers with ReLU activation~(Figure~\ref{fig:archi}(a)), CNN$_\text{MP}$ consists of three convolutional layers and max pooling layers followed by two fully connected layers~(Figure~\ref{fig:archi}(b)), CNN$_\text{None}$ is identical to CNN$_\text{MP}$ except for no max pooling layers, and CapsNet is that used in~\cite{sabour2017dynamic}. The four models were trained on MNIST and FashionMNIST, and their robustness was evaluated by the success rate of attack by the 5-SFIs generated using DE~(the number of candidate solutions $s$ is set to 500, and the iteration time $T$ is set to 1,000). Among the ten SFIs generated for each model and target class, the one with the highest confidence score is shown in Tables~\ref{tab:MNIST} and~\ref{tab:FashionMNIST}. Although all the models were fooled by the SFIs, CNN$_\text{MP}$ was particularly vulnerable to attack.

We consider the nature of the max pooling to be the cause of the vulnerability of CNN$_\text{MP}$. CNN$_\text{MP}$ first captures low-level features with a convolutional layer and a max pooling layer, and then applies successive layers. Finally, CNN$_\text{MP}$ fires the neurons that represent high-level features by connecting low-level features. However, the high-level features captured by CNN$_\text{MP}$ do not contain much spatial information of the low-level features because of max pooling. To classify objects with incomplete spatial information, CNN$_\text{MP}$ makes decisions based on whether lower-level features exist rather than how they are connected. The inappropriate processing of spatial information in CNN$_\text{MP}$ can also be responsible for the vulnerability of CNN$_\text{MP}$ to SFIs. Different pixels in the SFI can induce some neurons to fire through the convolution layers. The way the neurons fire by SFIs are patterns that would not be possible in natural images, but max pooling causes the CNN$_\text{MP}$ to map these inappropriate feature patterns to regions that are similar to those in natural images. 

MLP, CNN$_\text{None}$, and CapsNet, by contrast, connect low-level features to higher-level features without discarding the spatial information of the lower-level features. Their ability to represent the spatial connectivity between features more richly than CNN$_\text{MP}$ may require SFIs to fire neurons more appropriately, thus making it harder for SFIs to successfully attack. Although these robust architectures show a certain resistance against SFIs, it is difficult for MLP and CNN$_\text{None}$ to obtain a high image classification accuracy; in addition, the accuracy of CapsNet for large datasets such as ImageNet has yet to be studied owing to its high computational cost. To realize learning with a high classification accuracy and robustness against SFIs at the same time, a new DNN architecture that can handle large datasets without a loss of spatial information is required.

\begin{table*}[t]
  \caption{Comparison of four architectures trained on MNIST. Values in the table are the highest confidence scores~(\%) obtained by assessing ten SFIs generated by DE through the models. The three values represent differences in the base color and are in order of black, gray, and white. The accuracy and confidence scores next to the models are for the natural test dataset. Bold text represents values equal to or greater than average confidence score of natural dataset.}
  \centering
  \small
  \begin{tabular}{@{}cccccc@{}}
    \toprule
    model~(accuracy, confidence) & 0 & 1 & 2 & 3 & 4 \\ \midrule
    MLP~(98, 99) & 69/\textbf{99}/\textbf{100} & \textbf{100}/\textbf{99}/\textbf{99} & 55/\textbf{99}/0 & 13/\textbf{99}/1 & 91/\textbf{99}/0 \\
    CNN$_\text{MP}$~(99, 99) & \textbf{100}/\textbf{100}/\textbf{100} & \textbf{100}/\textbf{100}/\textbf{100} & \textbf{100}/\textbf{100}/\textbf{100} & \textbf{100}/\textbf{100}/\textbf{100} & \textbf{100}/\textbf{100}/\textbf{100} \\
    CNN$_\text{None}$~(98, 99) & \textbf{100}/0/0 & \textbf{100}/0/0 & \textbf{100}/\textbf{100}/\textbf{100} & \textbf{100}/0/0 & \textbf{100}/0/0 \\
    CapsNet~(99, 20) & 12/16/13 & 18/12/12 & 17/19/11 & 17/18/11 & 17/18/15 \\ \midrule\midrule
    & 5 & 6 & 7 & 8 & 9 \\ \midrule
    MLP~(98, 99) & \textbf{100}/97/0 & \textbf{99}/90/0 & \textbf{99}/\textbf{99}/0 & 50/93/68 & 97/97/0 \\
    CNN$_\text{MP}$~(99, 99) & \textbf{100}/\textbf{100}/\textbf{100} & \textbf{100}/\textbf{100}/\textbf{100} & \textbf{100}/\textbf{100}/0 & \textbf{99}/\textbf{99}/\textbf{100} & \textbf{100}/\textbf{100}/85 \\
    CNN$_\text{None}$~(98, 99) & \textbf{100}/0/0 & \textbf{100}/0/0 & \textbf{100}/0/0 & \textbf{100}/1/\textbf{99} & \textbf{100}/0/0 \\
    CapsNet~(99, 20) & \textbf{21}/18/9 & 12/16/12 & 18/18/12 & 14/\textbf{20}/12 & 14/17/14 \\
    \bottomrule
  \end{tabular}
  \label{tab:MNIST}
\end{table*}

\begin{table*}[t]
  \caption{Comparison in four trained models on Fashion-MNIST. The description of the table are the same as Table~\ref{tab:MNIST}}
  \centering
  \begin{tabular}{@{}cccccc@{}}
    \toprule
    model~(accuracy, confidence) & 0 & 1 & 2 & 3 & 4 \\ \midrule
    MLP~(89, 97) & 64/\textbf{100}/0 & 28/\textbf{100}/0 & 47/\textbf{99}/0 & 73/\textbf{99}/0 & 87/\textbf{98}/0 \\
    CNN$_\text{MP}$~(90, 99) & \textbf{100}/\textbf{100}/\textbf{100} & \textbf{100}/\textbf{100}/\textbf{100} & \textbf{100}/\textbf{100}/\textbf{100} & \textbf{100}/\textbf{100}/\textbf{100} & \textbf{100}/\textbf{100}/\textbf{100} \\
    CNN$_\text{None}$~(88, 94) & \textbf{99}/85/\textbf{99} & 78/\textbf{98}/\textbf{99} & 87/44/4 & \textbf{94}/52/0 & 93/67/24 \\
    CapsNet~(91, 20) & \textbf{21}/19/19 & 13/19/15 & 15/19/\textbf{20} & \textbf{20}/\textbf{20}/16 & 19/18/\textbf{20} \\ \midrule\midrule
    & 5 & 6 & 7 & 8 & 9 \\ \midrule
    MLP~(89, 97) & \textbf{100}/\textbf{100}/0 & 95/\textbf{100}/0 & \textbf{98}/\textbf{100}/0 & 27/\textbf{100}/\textbf{100} & 9/\textbf{100}/0 \\
    CNN$_\text{MP}$~(90, 99) & \textbf{100}/\textbf{100}/\textbf{100} & \textbf{100}/\textbf{100}/\textbf{100} & \textbf{100}/\textbf{100}/\textbf{100} & \textbf{100}/\textbf{100}/\textbf{100} & \textbf{100}/\textbf{100}/\textbf{100} \\
    CNN$_\text{None}$~(88, 94) & \textbf{100}/34/0 & \textbf{99}/70/20 & \textbf{100}/10/0 & \textbf{100}/\textbf{97}/\textbf{100} & 46/22/0 \\
    CapsNet~(91, 20) & \textbf{21}/18/15 & \textbf{21}/\textbf{20}/\textbf{20} & 18/18/17 & \textbf{22}/\textbf{21}/\textbf{21} & 18/17/15 \\
    \bottomrule
  \end{tabular}
  \label{tab:FashionMNIST}
\end{table*}

\begin{table}[t]
  \caption{Classification accuracy~(\%) of models trained on CIFAR10 and SFIs. The model classifies SFIs as the outlier class. Combined accuracy means the ratio of natural images and SFIs correctly classified. Natural accuracy refers to the ratio of correctly classified natural images. Defense accuracy refers to the ratio of correctly classified SFIs. Reattack accuracy refers to the ratio of successful attacks by the SFIs generated for model $M_i$.}
  \centering
  \small
  \begin{tabular}{@{}ccccc@{}}
    \toprule
    model     & combined & natural & defense & reattack \\ \midrule
    $M_0$     & None     & 80    & None    & 99      \\
    $M_1$     & 79     & 77    & 100    & 62      \\
    $M_2$     & 82     & 79    & 100    & 89      \\
    $M_3$     & 83     & 78    & 100    & 35      \\
    $M_4$     & 85     & 79    & 100    & 19      \\
    \bottomrule
  \end{tabular}
  \label{tab:detection2}
\end{table}

\begin{table}[t]
  \caption{Ratio of SFIs that successfully attacked each target class.}
  \centering
  \small
  \begin{tabular}{@{}cccccc@{}}
    \toprule
    model & plane    & car    & bird    & cat    & deer    \\ \midrule
    $M_0$ & \textbf{100} & \textbf{99} & \textbf{100} & \textbf{100} & \textbf{100} \\
    $M_1$ & 63 & 55 & 59 & \textbf{90} & 51 \\
    $M_2$ &\textbf{98}  & \textbf{90} & \textbf{97} & 82 & 47 \\
    $M_3$ & 78  & 42 & 38 & 29 & 0 \\
    $M_4$ & 2 & 75 & 1 & 19 & 27 \\ 
    $M_4$~(white) & \textbf{100}  & \textbf{99} & \textbf{100} & \textbf{100} & \textbf{100} \\ \midrule \midrule
    model & dog    & frog    & horse    & ship    & truck    \\ \midrule
    $M_0$ & \textbf{100} & \textbf{100} & \textbf{100}  & \textbf{100} & \textbf{100} \\
    $M_1$ &18 & 32 & 62  & \textbf{98} & \textbf{93} \\
    $M_2$ & \textbf{94} & \textbf{93} & \textbf{98}  & \textbf{96} & \textbf{95} \\
    $M_3$ & 13 & 0 & \textbf{93}  & 54 & 0 \\
    $M_4$ & 0 & 0 & 25  & 1 & 39 \\
    $M_4$~(white) & \textbf{100} & \textbf{98} & \textbf{100}  & \textbf{100} & \textbf{100} \\
    \bottomrule
  \end{tabular}
  \label{tab:detection3}
\end{table}

\subsection{SFI detection}
We also considered the defense of DNNs against SFIs. In particular, we used OOD detection for defense purposes. A typical method of OOD detection uses the outputs of the deep layers of the model~\cite{hendrycks2016baseline, lee2018simple}. However, these methods cannot be used because, as shown in Section~\ref{sec:pca}, natural images and SFIs share similar distributions; thus, it is difficult to distinguish them by using the dissociation of the distributions between natural images and SFIs. Here, starting from a VGG16 model $M_0$ trained using the CIFAR10 dataset, we repeatedly generated SFIs and retrained the model by incorporating these SFIs into the dataset. During the retraining, the model was trained to classify the SFIs into an outlier class. In the case of the CIFAR10 dataset, the outlier class becomes the 11th class. In the experiment, the following procedures were repeated for $i=0,\ldots,4$.

\begin{enumerate}
    \item 6,000 SFIs based on black~(denoted by $S_i$) are generated by DE for model $M_i$; 600 SFIs are generated for each class, and the number of changed pixels $k$ is sampled randomly from $[1, 100]$. It should be noted that SFIs are generated by targeting one of the 10 classes, and the retrained model is expected to classify them into the 11th class~(outlier class). The number of candidate solutions $s$ was set to 500, and the number of iterations $T$ was set to 3,000. When the confidence scores of a candidate solution in the group are higher than or equal to 90\% during DE, the SFI is determined by the candidate solution at that time~(early stopping). 
    \item Model $M_{i+1}$ is trained using CIFAR10 and 5,000 SFIs of $S_j, (j=0,\ldots,i)$. Unused 1,000 SFIs of $S_j$ will be used for test.
\end{enumerate}

This approach results in an imbalance in the number of data points between the SFI dataset and each class of CIFAR10~(much larger number of SFIs). This imbalance makes it more difficult for SFIs to fool the model. The results of the training are listed in Table~\ref{tab:detection2}. Each model perfectly classified the prepared test dataset of the SFIs as the outlier class without losing the accuracy on the natural test dataset. This seems to be intuitively correct because SFIs have a clearly different distribution from natural images in input space. However, new SFIs can still be generated to attack these models. Table~\ref{tab:detection3} shows the ratio of the SFIs that successfully attacked each target class. Even for model $M_4$, which has the highest detection accuracy, the SFIs targeting the car class succeed in attacking at a high rate. We also attacked the model $M_4$ with white-based SFIs to determine the extent to which OOD detection depends on the base color. The generated SFIs were found to fool model $M_4$ at a rate of 99\%~(Table~\ref{tab:detection3}). This indicates that models trained with SFIs are still vulnerable to SFIs in different base colors that are not used during training. Note that SFIs can be removed using a simple filtering, but SFIs are only a minimal case of images that are extremely far from natural images. Filtering all such cases is difficult and SFIs could become a potential risk in most DNN-based systems.

\subsection{Transferability}
\label{sec:trans}
Here, we observe the transferability of SFIs on CIFAR10. We input 5-SFIs generated in the source model to the destination model and counting SFIs which yield high confidence scores~($\geq$ 90\%) in both models. These results are shown in Table~\ref{tab:transferability}. Most SFIs are not able to attack other models, but some SFIs produce high confidence scores in the two models. Also, SFIs generated in a robust model tend to succeed in attacking another more vulnerable model. In particular, 45\% of SFIs generated in the ResNet50 model fooled the VGG16 model with high confidence scores. Ilyas~\etal~have hypothesized that the success of adversarial attacks is attributed to non-robust features in a training dataset~\cite{ilyas2019adversarial}. Non-robust features are imperceptible to humans and improve the generalized performance of DNNs. They have insisted that adversarial attacks add perturbation like non-robust features to natural images and fool DNNs. Therefore, according to this theory, the transferability of adversarial attacks can be explained by non-robust features, which are commonly learned by several models. Altered pixels of SFIs might be such non-robust features, so SFIs have transferability. 

\begin{table}[t]
  \caption{Result of transferability. The values in this table are the number of SFIs and the ratio, which are calculated by inputting 5-SFIs generated in source model to destination model and counting SFIs which yield high confidence scores~($\geq$ 90\%) in both models.}
  \centering
  \begin{tabular}{@{}ccccc@{}}
    \toprule
     & & \multicolumn{3}{c}{destination}\\\cmidrule(r){3-5}
     & & VGG16 & NiN & ResNet50 \\ \midrule
    \multirow{3}{*}{\!\!\!source\!\!\!\!\!}&VGG16 & \!\!1947~(1.00)\!\! & \!\!202~(0.10)\!\! & \!\!188~(0.09)\!\! \\
    &NiN & \!\!338~(0.28)\!\! & \!\!1170~(1.00)\!\! & \!\!110~(0.09)\!\! \\
    &ResNet50 & \!\!309~(0.45)\!\! & \!\!213~(0.31) \!\!& \!\!672~(1.00)\!\! \\
    \bottomrule
  \end{tabular}
  \label{tab:transferability}
\end{table}

\subsection{Relation to Adversarial Attacks}
\label{sec:relation}
In this paper, we do not conduct attack efficiency comparison to standard adversarial attacks~\cite{szegedy2013intriguing, carlini2017towards, chen2017ead, dong2018boosting, moosavi2016deepfool,eykholt2018robust, sharif2016accessorize, sharif2019general, thys2019fooling, goodfellow2014explaining, Croce_2019_ICCV, Modas_2019_CVPR, narodytska2016simple, papernot2016limitations} and the one pixel attack~\cite{su2019one}. The concept of our paper is to show images that are extremely far from natural images can also fool CNNs. We are not claiming that SFIs are more efficient than the other attack methods. 

One may think that our SFIs are conceptually similar to the one pixel attack~\cite{su2019one}. However, as we discussed in~\ref{sec:related}, the one pixel attack adds one pixel noise to {\it natural} images, whereas our SFIs are showing new potential threats that featureless images can also fool CNNs. Besides, theoretical analysis of SFIs is novel as compared to~\cite{su2019one}.

\section{Conclusion}
We addressed and affirmatively answered the fundamental question of whether it is possible to fool DNNs by images that are extremely unrecognizable to humans. We proposed SFIs as a minimal case of such images and revealed the vulnerability of DNNs to SFIs. SFIs are not recognizable to humans as natural objects, but are classified by a DNN with high confidence scores. In contrast to existing fooling images, SFIs have neither local nor global features. We discussed the conditions for the existence of SFIs in three linear and nonlinear settings. This theoretical analysis indicates that complex models are more susceptible to SFI attacks. For empirical analysis, we proposed two methods for SFI generation, one based on DE in a class-targeted setting, and the other inverted the deep features of a natural image under an image-targeted setting. In the experiments, we demonstrated that SFIs and natural images become quite similar to each other in the deep layers of the DNN, although at the input domain, they are clearly different in appearance. This explains why SFIs can fool complex DNNs. We compared four architectures~(MLP, CNN with max pooling layers, CNN without them, and CapsNet) regarding their robustness against SFIs and showed that a CNN$_\text{MP}$ is more vulnerable to SFIs than other architectures. We also observed the vulnerability of max pooling, which loses spatial information of low-level features. To develop a defense against SFIs, we used OOD detection, which is able to detect the SFIs used for the test dataset with 100\% accuracy. However, it was possible to generate further SFIs that can attack the retrained model. Lastly, we showed SFIs have the transferability as well as adversarial examples. This implicates that the altered pixels of SFIs are unrecognizable but commonly learned by different models.

In a future study, since SFIs are minimal cases of images that are extremely different from natural images, we further analyze the vulnerability of DNNs by finding other such images. Furthermore, as some studies have shown that adversarial examples can be used for data augmentation to improve clean accuracy~\cite{kim2020m2m,xie2020adversarial}, we consider that SFIs can be used as one of the augmentation methods and improve the clean accuracy of the test set.

\bibliographystyle{ieeetr}
\bibliography{main}

\end{document}